\documentclass[journal]{IEEEtran}
\usepackage{amsmath,amsfonts}
\usepackage{algorithmic}
\usepackage{array}
\usepackage[caption=false,font=normalsize,labelfont=sf,textfont=sf]{subfig}
\usepackage{textcomp}
\usepackage{stfloats}
\usepackage{url}
\usepackage{verbatim}
\usepackage{graphicx}
\usepackage{hyperref}
\def\BibTeX{{\rm B\kern-.05em{\sc i\kern-.025em b}\kern-.08em
    T\kern-.1667em\lower.7ex\hbox{E}\kern-.125emX}}
\usepackage{balance}

\usepackage{color,soul}

\begin{document}
\title{
LATS: Large Language Model Assisted Teacher-Student Framework for Multi-Agent Reinforcement Learning in Traffic Signal Control
}
\author{Yifeng Zhang, Peizhuo Li, Tingguang Zhou, Mingfeng Fan, Guillaume Sartoretti
\thanks{This work is supported by A*STAR, CISCO Systems (USA) Pte. Ltd and National University of Singapore under its Cisco-NUS Accelerated Digital Economy Corporate Laboratory (Award I21001E0002). (\textit{Corresponding author: Mingfeng Fan}).}
\thanks{Yifeng Zhang, Peizhuo Li, Mingfeng Fan, and Guillaume Sartoretti are with the Department of Mechanical Engineering, National University of Singapore (E-mail: yifeng@u.nus.edu, e0376963@u.nus.edu, e1192634@u.nus.edu, ming.fan@nus.edu.sg, guillaume.sartoretti@nus.edu.sg)}
}

\maketitle

\begin{abstract}
Adaptive Traffic Signal Control (ATSC) aims to optimize traffic flow and minimize delays by adjusting traffic lights in real time. 
Recent advances in Multi-agent Reinforcement Learning (MARL) have shown promise for ATSC, yet existing approaches still suffer from limited representational capacity, often leading to suboptimal performance and poor generalization in complex and dynamic traffic environments.
On the other hand, Large Language Models (LLMs) excel at semantic representation, reasoning, and analysis, yet their propensity for hallucination and slow inference speeds often hinder their direct application to decision-making tasks. 
To address these challenges, we propose a novel learning paradigm named LATS that integrates LLMs and MARL, leveraging the former's strong prior knowledge and inductive abilities to enhance the latter's decision-making process. 
Specifically, we introduce a plug-and-play teacher-student learning module, where a trained embedding LLM serves as a teacher to generate rich semantic features that capture each intersection's topology structures and traffic dynamics. 
A much simpler (student) neural network then learns to emulate these features through knowledge distillation in the latent space, enabling the final model to operate independently from the LLM for downstream use in the RL decision-making process. 
This integration significantly enhances the overall model's representational capacity across diverse traffic scenarios, thus leading to more efficient and generalizable control strategies. 
Extensive experiments across diverse traffic datasets empirically demonstrate that our method enhances the representation learning capability of RL models, thereby leading to improved overall performance and generalization over both traditional RL and LLM-only approaches.
These results demonstrate our framework's effectiveness across a wide range of complex ATSC tasks, opening up a new avenue for the integration of RL and LLMs in practical traffic signal control applications.
\end{abstract}


\section{Introduction}

As the number of vehicles in cities continues to grow, Traffic Signal Control (TSC) systems play an increasingly important role in maintaining an orderly traffic flow, ensuring safety, and improving efficiency and accessibility in ever-expanding urban environments.
Efficient TSC systems offer a cost-effective solution that minimizes the need for extensive modifications to existing infrastructure while reducing commuting times, easing congestion, lowering vehicle emissions, and significantly enhancing urban living conditions~\cite{wei2019survey, haydari2020deep}.
However, commonly used fixed-time and rule-based signal control systems often fail to adapt to complex and dynamic traffic conditions. 
To address this, various Adaptive Traffic Signal Control (ATSC) systems like SCOOT~\cite{hunt1982scoot} and SCATS~\cite{pr1992scats} have been developed to optimize traffic flow and reduce congestion by adjusting signals in real time based on current traffic demands.

\begin{figure}[t]
    \centering
    \includegraphics[
    width=\linewidth]
    {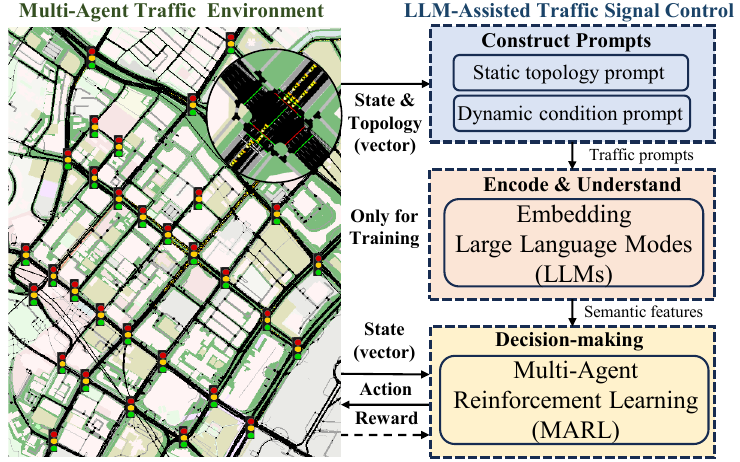}
    \caption{The pipeline of our proposed LLM-assisted TSC framework, where we employ LLMs to generate rich semantic features based on customized traffic prompts to enhance the downstream MARL decision-making process.}
    \label{fig:intro}
\end{figure}

Recent advances in Multi-Agent Reinforcement Learning (MARL) approaches have shown substantial progress in developing adaptive signal control strategies.
Existing prior RL-based work primarily focuses on three key areas: (1) Enhancing RL agent design, i.e., the definitions of state, action, and reward spaces for TSC tasks~\cite{zhang2022expression, liang2022oam, wei2019presslight, zheng2019diagnosing, wang2024traffic}. 
(2) Facilitating efficient collaboration and cooperation among agents~\cite{chu2019multi, wei2019colight, goel2023sociallight, liu2023gplight}. 
(3) Improving generalization across different scenarios~\cite{zheng2019learning, zang2020metalight, oroojlooy2020attendlight, zhu2023metavim, jiang2024general}. 
These methods often rely on parameter-sharing mechanisms to improve learning efficiency and stabilize multi-agent training.
However, the limited representational capacity of small, parameter-sharing RL models becomes more pronounced in network-wide ATSC, where diverse intersection topologies and dynamic traffic demands make it challenging to learn robust and effective control strategies, often resulting in suboptimal performance.

On the other hand, Large Language Models (LLMs) have shown promising results in various downstream tasks such as traffic prediction~\cite{liu2024spatial,li2024urbangpt}, robotics~\cite{ahn2022can, shah2023lm}, and autonomous driving~\cite{wen2023dilu, fu2024drive} due to their advanced abilities in inductive reasoning and understanding contextual information. 
However, LLMs are prone to ``hallucination,'' where the model generates plausible but factually incorrect information. 
To make matters worse, as the length of input prompts increases, the inference time and computational demands of LLMs also increase. 
These factors make it difficult to directly apply LLMs in real-world TSC systems to ensure robust and responsive control over complex, dynamic traffic flows.
Nevertheless, LLMs provide strong semantic and contextual understanding that can substantially enhance the representational capacity of RL models in diverse and complex traffic scenarios.

To address these challenges, we introduce \textbf{LATS}, a novel \textbf{L}arge Language Model \textbf{A}ssisted \textbf{T}eacher-\textbf{S}tudent learning framework for multi-agent traffic signal control (MATSC). 
As illustrated in Fig.~\ref{fig:intro}, we leverage the exceptional prior knowledge and contextual understanding abilities of LLMs to process unstructured traffic semantic data, integrating their outputs to enhance the model's representational ability across diverse traffic scenarios.
Specifically, we introduce a plug-and-play teacher-student learning module, where a trained embedding LLM serves as a knowledgeable teacher during training, generating rich semantic features from traffic prompts that capture each intersection’s topology and traffic dynamics.
Embedding LLMs are less prone to hallucinations because they focus on accurate semantic representation rather than generating creative open-ended text.
The student model learns to emulate these semantic features through knowledge distillation in the latent space. 
This is achieved by applying two variational autoencoders (VAEs), which encode the semantic features from the teacher LLM and the state vectors from the student model into a shared latent space, aligning their latent distributions by minimizing the Kullback–Leibler (KL) divergence.
We then integrate the student's latent features into an RL-based general TSC framework to enhance its phase representations for the downstream RL decision-making process.
By doing so, we transfer the semantic representations from the LLM to the student during training, enhancing the RL model’s representational capacity for more effective policy learning.
Thus, the agent can operate independently without the LLM during execution, mitigating hallucination risks and avoiding the computational overhead of online LLM calls.

We conduct comprehensive simulation experiments using open-source traffic datasets, which include a homogeneous Grid network with 25 identical intersections and a realistic heterogeneous Monaco network with 28 distinct intersections.
Our results empirically show that LATS outperforms other baseline methods, including traditional, pure LLM-based, and pure RL-based methods, across various traffic optimization metrics.
Additionally, we show that our method exhibits superior zero-shot transferability to diverse spatio-temporal traffic demands compared to other RL-based methods, as well as greater robustness and reduced inference time compared to pure LLM-based methods.
These results underscore the effectiveness of LATS, where semantic features distilled from LLMs significantly enhance the model's representation capability, thus leading to more generalizable and efficient signal control strategies.
Finally, we show that integrating our LLM-assisted teacher-student learning framework into other pure RL-based approaches further enhances their performance by leveraging the richer semantic representations.

In summary, our main contributions are summarized as:

\begin{itemize}
    \item We propose \textbf{LATS}, a novel LLM-assisted MARL framework for network-wide ATSC that leverages an LLM to generate semantic representations of intersection topology and traffic dynamics, which are integrated into the RL training process, thereby improving the model’s representational capacity and enabling more effective and generalizable control strategies. To the best of our knowledge, this is the first work to leverage the semantic representation capability of LLMs for RL-based ATSC.
    \item We introduce a plug-and-play teacher–student distillation module, where a teacher LLM encodes prompts into semantic embeddings during training, and a much simpler student network learns to emulate them in the latent space.
    This process distills the semantic representations generated by the LLM into the RL model, enabling it to better capture the structural and dynamic characteristics of traffic environments for downstream ATSC tasks, while supporting LLM-free inference during deployment.
    \item We conduct extensive experiments on open-source traffic datasets, including both homogeneous and heterogeneous networks, demonstrating the effectiveness and transferability of our LATS compared with various baselines, and showing its ability to enhance the performance of existing RL-based TSC methods when integrated with them.
\end{itemize}

\section{Related Works}

\subsection{RL-based ATSC}
Recent advances in MARL demonstrate significant potential in ATSC tasks, which can be categorized into three areas: RL agent design, collaboration and cooperation, and general policy learning.
First, \textbf{RL Agent Design} focuses on developing effective state and reward definitions to enhance the RL agents' ability to learn and adapt to dynamic traffic conditions. 
For instance, max-pressure (MP)~\cite{wei2019presslight}, advanced traffic state (ATS)~\cite{zhang2022expression}, cell transmission state (CTS)~\cite{liang2022oam}, and queue dynamic state encoding (QDSE)~\cite{zhang2025coordlight} are proposed to represent local traffic conditions better. 
Additionally, metrics such as queue length~\cite{zheng2019diagnosing, chu2019multi}, pressure~\cite{wei2019presslight}, regularized delay~\cite{liang2022oam}, and IFDG~\cite{lin2023denselight} are proposed as effective surrogate dense optimization objectives.
Second, \textbf{Collaboration and Cooperation} focuses on designing advanced MARL strategies that enable agents to optimize traffic flow and reduce congestion jointly.
Some works rely on advanced communication or feature aggregation mechanisms, such as graph neural networks~\cite{wei2019colight, wang2020stmarl, devailly2021ig, zeng2021graphlight} and attention networks~\cite{wu2021dynstgat, mao2022mastering, zhang2025coordlight}, to facilitate collaboration among agents. 
Others utilize centralized training with decentralized execution (CTDE) frameworks~\cite{chen2021collaborative, goel2023sociallight, liu2023gplight, zhou2024cooperative} to achieve effective cooperation by addressing credit assignment problems in optimizing joint objectives.
Finally, \textbf{General Policy Learning} aims to develop control strategies that can be applied to different traffic networks and demands, thereby improving their generalization and transferability to new scenarios. 
This line of work can be divided into two directions.
The first direction~\cite{liang2022oam, zheng2019learning, oroojlooy2020attendlight, jiang2024general, wang2024unitsa, zhang2024HeteroLight, qiao2024hol} develops \textbf{general ATSC frameworks} by unifying intersection topologies, allowing parameter-sharing models to extract features and learn policies across both homogeneous and heterogeneous networks. 
The second direction~\cite{zang2020metalight, zhu2023metavim, jiang2024general, wang2022meta, lu2023dualight, zhang2020generalight, jiang2024x} focuses on \textbf{meta-learning and cross-scenario methods} that enhance the scalability and transferability of RL approaches to unseen traffic environments.
Overall, most RL-based ATSC methods adopt parameter sharing to improve scalability and training efficiency in large-scale traffic networks. 
However, their limited model capacity and single-modal inputs still constrain representation learning and generalization, making it difficult to learn effective and robust control strategies under complex and dynamic traffic conditions.
To address these challenges, we propose the LATS framework, which introduces LLM-assisted semantic representation learning to enhance the model’s representational capacity and improve control performance in complex and dynamic traffic scenarios.

\subsection{LLM-based ATSC}

LLMs have recently shown remarkable potential in diverse control tasks due to their zero-shot reasoning and generalization capabilities~\cite{ahn2022can, shah2023lm, wen2023dilu, driess2023palm, da2024open, wang2025large, sun2024llm}.
Building on this, recent studies have explored the use of LLMs as teachers to guide RL or decision-making processes.
For instance, Zhou et al.~\cite{zhou2023large} proposed LLM4Teach, where an LLM guides a student agent within RL environments. Through knowledge distillation, the student eventually outperforms the teacher on benchmarks such as MiniGrid and Habitat.
Xu et al.~\cite{xu2025tell} introduced TeLL-Drive, where an LLM teacher with chain-of-thought reasoning assists the training of autonomous driving policies, leading to more efficient and robust learning.
Baumann et al.~\cite{baumann2025enhancing} presented a hybrid approach that integrates an LLM with a model predictive controller (MPC), in which the LLM provides high-level reasoning and the MPC ensures safety through reliable low-level execution.
Beyond single-agent settings, several studies have applied LLMs to cooperative driving in Connected and Autonomous Vehicles (CAVs)~\cite{liu2025language, fang2025towards, hu2024agentscodriver}. 
For example, Liu et al.~\cite{liu2025language} proposed LDPD, where an LLM teacher provides demonstrations to guide MARL agents through policy distillation. 
Together, these studies highlight the feasibility of LLMs serving as teachers in practical learning or control processes.
Yet, their potential in learning and leveraging rich semantic representations for downstream decision-making remains underexplored.

In parallel, knowledge distillation (KD) for LLMs has emerged as a rapidly growing research direction~\cite{xu2024survey, yang2024survey, di2024performance, tan2023gkd}.
Xu et al.~\cite{xu2024survey} conducted a comprehensive survey of KD techniques for LLMs, organizing them into three main pillars: algorithm (the design of distillation methods), skill (the transfer of task-specific abilities), and verticalization (adaptation to specific domains or applications).
These methods demonstrate that LLMs can effectively serve as teachers. 
The distilled student models retain essential capabilities while remaining lightweight and efficient, which makes LLM-based distillation a practical solution for real-world deployment.

Following these advances, researchers have also started to investigate the application of LLMs in ATSC.
LLMLight~\cite{lai2023large} employed LLMs as decision-making agents for ATSC by converting real-time traffic conditions into traffic prompts and leveraging their reasoning abilities to generate control strategies.
LA-Light~\cite{wang2024llm} presented a hybrid TSC framework that enhances LLMs' decision-making ability with a suite of perception and decision-making tools, enabling them to interpret and respond to complex traffic scenarios with rare and unexpected events.
To mitigate the simulation-to-real gap,~\cite{da2024prompt} utilized LLMs to profile and reason about system dynamics, which enables RL agents to learn realistic policies by understanding traffic dynamics through prompt-based action transformation. Tang et al.~\cite{tang2024large} further investigated the application of LLMs in arterial TSC, showing great improvements in complex urban environments.
Despite these efforts to employ LLMs for ATSC tasks, most validations are conducted in simple, homogeneous traffic networks, and their decision-making processes still face significant challenges due to the inherent hallucinations and the substantial computational costs.

\section{Preliminaries}
In this section, we first define the key terminologies for the MATSC problem, then present its formulation as a MARL problem, and finally describe the design of the RL agent, including the state, action, and reward definitions.

\subsection{Traffic Terminology}
\textbf{\textit{Definition 1 (Road and Lane)}}: A traffic road is a designated pathway for vehicles, typically consisting of multiple lanes. A lane is a subdivision of a road that directs and organizes traffic, allowing vehicles to travel in an orderly manner.

\textbf{\textit{Definition 2 (Traffic Movement):}} A specific path that vehicles follow to navigate through an intersection, moving from an incoming lane (entry point) to a connected outgoing lane (exit point). The traffic movement $m_{ij} \in \mathcal{M}$ between incoming lane $i$ and outgoing lane $j$ is defined as $m_{ij}$.

\textbf{\textit{Definition 3 (Traffic Signal Phase):}} 
Each phase controls the activation status of traffic movements, with each traffic movement defined as activated ($m_{ij}=1$) to permit passage, or deactivated ($m_{ij}=0$) to prohibit passage.
Let $\mathcal{M}$ represent the complete set of movements at the intersection, and $\mathcal{M}_{p} \in \mathcal{M}$ the subset of non-conflicting activated movements for phase $p$. Then, the phase $p$ can be formally defined as: $p=\{m=1 \mid m \in \mathcal{M}_p\} \cup \{m=0 \mid m \in \mathcal{M} \setminus \mathcal{M}_p\}$.

\textbf{\textit{Definition 4 (Traffic Agent and Network):}} A traffic agent manages the traffic phases or timings according to real-time traffic conditions at the intersection. 
Additionally, a traffic network is a multi-agent system that links various traffic agents via interconnected roads.
Homogeneous traffic networks consist of intersections/agents with identical topology structures, while heterogeneous traffic networks feature intersections/agents with diverse topology structures.

\begin{figure}[t]
    \centering
    \includegraphics[
    width=\linewidth]{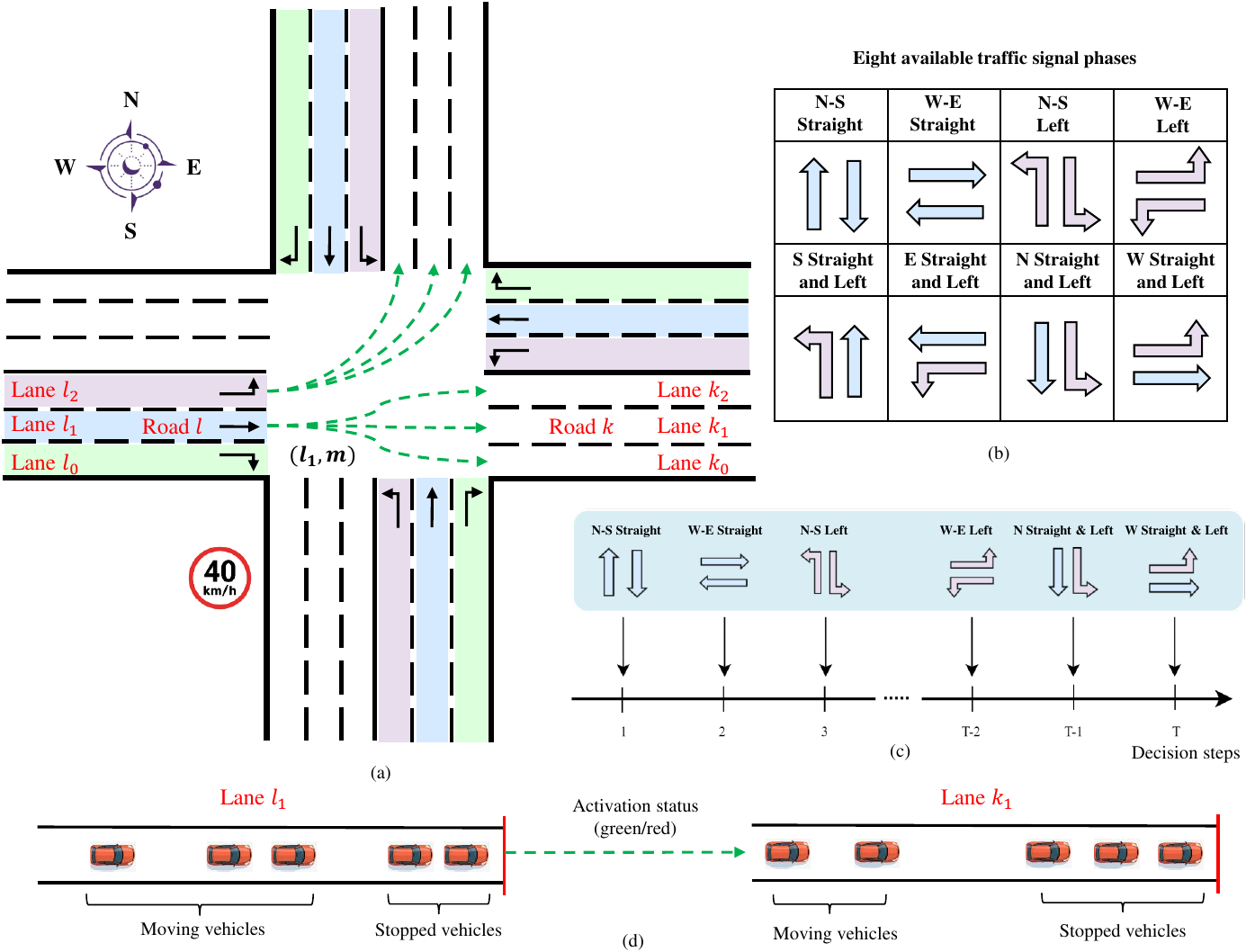}
    \caption{(a) An illustration of an intersection, which consists of roads, lanes, and traffic movements. (b) Eight non-conflicting traffic signal phases at the intersection. (c) Depiction of the traffic signal control process based on phase selections. (d) Illustration of the state of a traffic movement for RL agents.}
    \label{fig:intersection}
\end{figure}

In Fig.~\ref{fig:intersection}(a), we show a typical 4-arm intersection with four incoming roads and four outgoing roads, such as road $l$ and road $k$.
Each road has three lanes, leading to twelve incoming lanes (e.g., lane $l_0$, lane $l_1$, and lane $l_2$) and twelve outgoing lanes (e.g., lane $k_0$, lane $k_1$, and lane $k_2$).
Since each incoming lane can connect to multiple outgoing lanes, there are a total of thirty-six traffic movements within the intersection.
We define eight different traffic phases for this intersection by grouping non-conflicting movements, as illustrated in Fig.~\ref{fig:intersection}(b). Following the settings of previous work~\cite{chu2019multi, wei2019colight, goel2023sociallight}, we define the ATSC problem as selecting the next phase to be activated for a predetermined duration, as shown in Fig.~\ref{fig:intersection}(c). 
We set the phase duration between two phase selections to \textbf{10 seconds}. Changing the phase triggers a \textbf{3-second} yellow phase, followed by \textbf{7 seconds} of the selected new phase.
Furthermore, Fig.~\ref{fig:intersection}(d) illustrates the state of a traffic movement connecting incoming lane \( l_1 \) and outgoing lane \( k_1 \). This state includes the movement's activation status and the number of moving and stopped vehicles measured by 50-meter lane-area detectors in both lanes, which are used to calculate the state vector of the intersection.

\subsection{Problem Formulation}
We formulate the MATSC problem as a MARL problem, where each signalized intersection is considered as a learning agent. This problem is modeled as a Decentralized Partially Observable Markov Decision Process (Dec-POMDP)~\cite{oliehoek2016concise}, represented by a tuple $G=<\mathcal{N}, \mathcal{S}, \mathcal{A}, \mathcal{R}, \mathcal{T}, \mathcal{O}, \mathcal{Z}, \rho_0, \gamma>$.
Here, $\mathcal{N}=\{1,\cdots,n\}$ is the set of traffic agents. $\mathcal{S}$ is the global state space and $\mathcal{A}=\bigcup_{i=1}^{n} \mathcal{A}_{i}$ is the joint action space.
At each decision time step $t$, each agent $i$ selects an action $a_{i, t} \in \mathcal{A}_{i}$, contributing to a joint action $\mathbf{a}_t=\left\{a_{i, t}\right\}_{i=1}^{k_t} \in \mathcal{A}$, which causes an environmental state transition according to $\mathcal{P}\left(s_{t+1} \mid s_t, \mathbf{a}_t\right):\mathcal{S}\times\mathcal{A}\times\mathcal{S} \rightarrow[0,1]$. Each agent then receives a reward from its reward function $r_{i, t}=\mathcal{R}_i\left(s_t, \mathbf{a}_t\right):\mathcal{S}\times\mathcal{A}\rightarrow\mathbb{R}$. The initial state distribution is given by $\rho_0$ and $\gamma$ is the discount factor. 
In a partially observable setting, agents cannot directly access the full state of the environment $s \in \mathcal{S}$. Instead, each agent draws an individual observation $o_{i,t}$ based on the observation function $o_{i, t}=\mathcal{Z}_i\left(s_t, a_t\right): \mathcal{S}\times\mathcal{A} \rightarrow \mathcal{O}$. 
We also define the episodic trajectory $\tau$ as $\tau=\left\{\left(s_t, \mathbf{a}_t, r_{1, t}, \ldots, r_{n, t}\right)\right\}_{t=0}^T$, where $T$ represents the maximum environment step.
Given policy $\pi_i$ parameterized by $\theta_i$, the objective is to find the optimal policy $\pi^{*}$ that maximizes the discounted sum of rewards for all agents: $J\left(\theta_i\right)=\mathbb{E}_\tau\left[R_i(\tau)\right]$, where $R_i(\tau)=\sum_{t=0}^{T} \gamma^{t} r_{i, t}$.

\begin{figure*}[t]
    \centering
    \includegraphics[width=\linewidth]{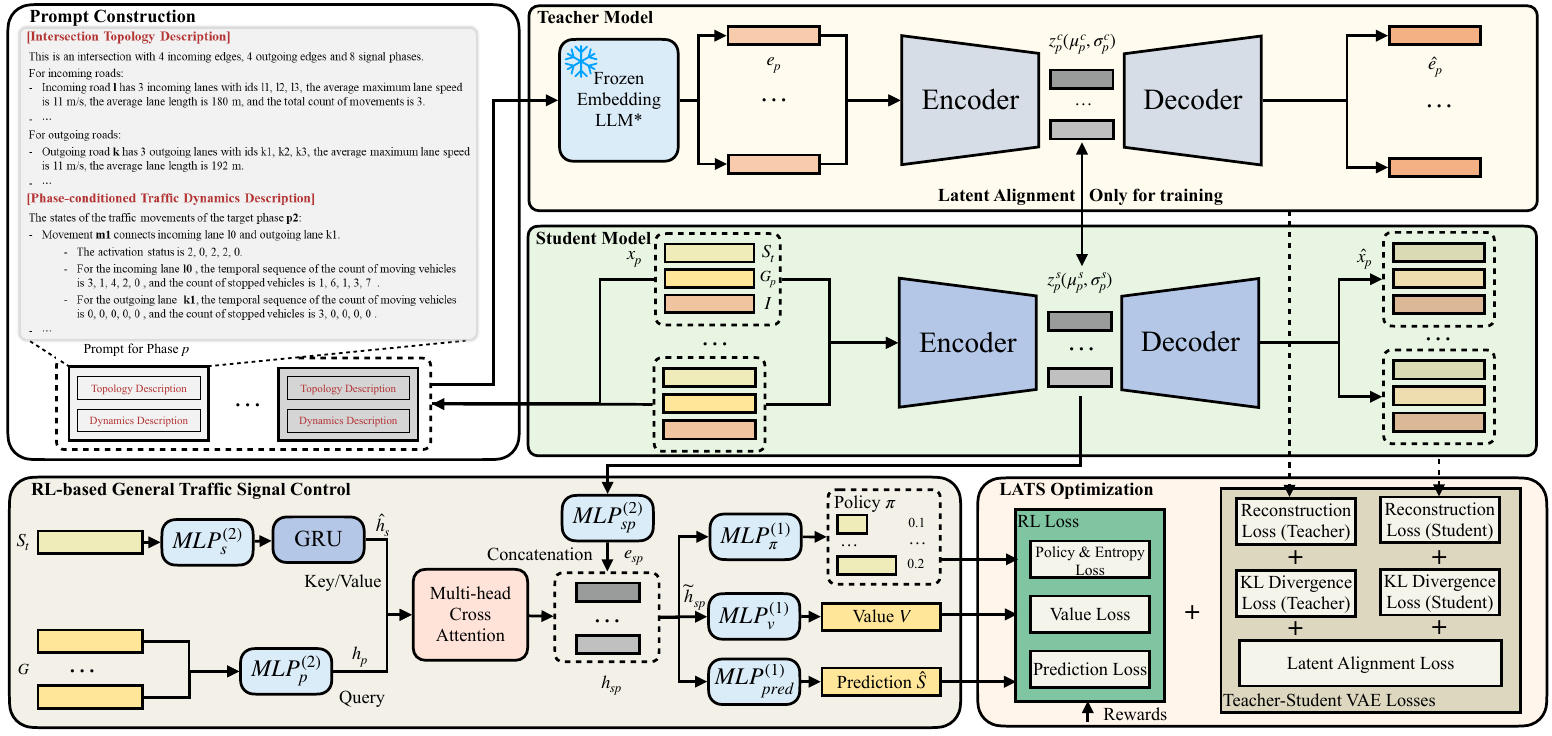}
    \caption{Architecture of the proposed LLM-assisted teacher-student learning framework (LATS) for general TSC. The teacher uses a pre-trained embedding LLM to generate semantic embedding vectors from customized traffic prompts (top left) for each phase. The student model emulates these vectors through knowledge distillation in the latent space. These latent vectors are integrated into the downstream RL decision-making process to enhance the phase representations.}
    \label{fig:method}
\end{figure*}

\subsection{RL Agent Design}
We consider a general RL framework that is applicable to both homogeneous and heterogeneous networks; thus we define the local state, action, and reward space as follows:

\subsubsection{Local state/Observation}
We define the state vector for a single traffic movement $m_{(l_{in}, l_{out})}$ at an intersection as a combination of five lane features vectors at time step $t$ as:
\begin{equation}
   S_{m}(t) \in \mathbb{R}^{5} =[P_{m}(t), Q_{{in}}(t), Q_{{out}}(t), N_{{in}}(t), N_{{out}}(t)],
\end{equation}
where $P_{m}(t)\in\{0,1\}$ represents the activation status of the current movement $m$ (1 if the movement is active, 0 otherwise), $Q_{{in}}(t)$ and $Q_{{out}}(t)$ are the count of stopped vehicles on the incoming and outgoing lanes of $m$, respectively, and $N_{{in}}(t)$ and $N_{{out}}(t)$ refer to the number of moving vehicles on these lanes.
To ensure consistency across intersections, all four vehicle-related features are represented as scalars normalized to the range $[0,1]$, measured by 50-meter lane-area detectors installed near the intersection.
Thus, the time-variant local state vector for a single phase $p$ at the intersection is expressed as:
\begin{equation}
\label{state_function}
    S_{p,t} = S_{p}(t) \in \mathbb{R}^{|\mathcal{M}_p| \times 5}=[S_m(t) \mid m \in \mathcal{M}_p].
\end{equation}
The state vector for a single intersection is written as:
\begin{equation}
    S_t = S(t) \in \mathbb{R}^{|\mathcal{M}| \times 5}=[S_m(t) \mid m \in \mathcal{M}], 
\end{equation}
which includes the states of all available traffic movements. Additionally, we define the time-invariant phase state vector used to indicate the activation status of traffic movements for a given traffic phase $p$ at the intersection as:
\begin{equation}
G_p \in \mathbb{R}^{|\mathcal{M}|} =[1 \text { if } m \in \mathcal{M}_{p} \text { else } 0 \mid m \in \mathcal{M}],
\end{equation}
Given the total phase set $\mathcal{P}$ of an intersection, we construct the comprehensive phase state vector $G$ as: 
\begin{equation}
G \in \mathbb{R}^{|\mathcal{P}| \times |\mathcal{M}|} = [G_p \mid p \, \in \, \mathcal{P}].
\end{equation}
Additionally, we formulate the time-invariant intersection topology vector for an intersection in the following manner: 
\begin{equation}
I = [T_{\text{type}}, L^{\mathrm{in}}, V_{\text{max}}^{\mathrm{in}}, N_{\text{l}}^{\mathrm{in}}, N_{\text{m}}^{\mathrm{in}}, L^{\mathrm{out}}, V_{\text{max}}^{\mathrm{out}}, N_{\text{l}}^{\mathrm{out}}],
\end{equation}
where $T_{\text{type}}$, encoded as a one-hot vector, specifies the phase setting type of the intersection (e.g., 3-phase, 4-phase, or 5-phase). 
$L^{\mathrm{in}}$, $V_{\text{max}}^{in}$, $N_{\text{l}}^{in}$, and $N_{\text{m}}^{in}$ represent the average lane length, maximum speed, number of lanes, and total traffic movements on incoming roads, respectively. Similar metrics for outgoing roads include $L_{\mathrm{out}}$, $V_{\text{max}}^{out}$, and $N_{\text{l}}^{out}$.

\subsubsection{Action}
Consistent with prior works~\cite{zhang2022expression, wei2019presslight, chu2019multi, wei2019colight, goel2023sociallight, jiang2024general}, we define the action space for each agent as a finite set of collision-free traffic phases, also denoted as $\mathcal{P}$. 
Thus, agents synchronously select and implement a traffic phase from this set for a predetermined duration.

\subsubsection{Reward}
The reward for each agent is defined as the negative sum of stopped vehicle counts, measured by lane-area detectors with a 50-meter detection range on the incoming lanes near the intersection.
Given the set of incoming lanes at the intersection $\mathcal{L}_{in}$, the reward is calculated as: $R(t) = - \sum_{l_{in} \in \mathcal{L}_{in}} {Q}_{l_{in}}(t)$, where ${Q}_{l_{in}}(t)$ is the stopped vehicle count measured on each incoming lane $l_{in}$.

\section{LLM-assisted Teacher-Student TSC}
In this section, we introduce the proposed LATS framework, illustrated in Fig.~\ref{fig:method}, which consists of an RL-based TSC module and an LLM-assisted teacher-student module. 
The teacher network, which incorporates a pre-trained LLM, is used only during training to generate semantic representations for distillation, while a lightweight student network learns to emulate these representations and integrates them into the RL-based TSC module.
Unlike conventional knowledge distillation for policy imitation or guidance, our method distills semantic representations from the LLM to strengthen the RL model’s feature representations.
During execution, LATS relies solely on the RL-based TSC and the student network, without the need for the LLM. 
The overall objective combines RL loss, VAE losses, and latent alignment loss, enabling joint optimization of both modules. 
Notably, the teacher-student module is designed as a plug-and-play component, making it compatible with various ATSC methods to enhance control performance and generalization across complex traffic environments.

\subsection{RL-based TSC}
In this section, we present our RL-based general TSC framework for MATSC, aiming to learn a parameter-sharing policy for intersections with diverse topologies, as shown at the bottom of Fig.~\ref{fig:method}.
Here, we adopt the general TSC framework proposed in HeteroLight~\cite{zhang2024HeteroLight} due to its simplicity and efficiency. 
However, our approach is not limited to this RL paradigm; it is flexible enough to be applied to other general TSC methods, such as FRAP~\cite{zheng2019learning} and AttendLight~\cite{oroojlooy2020attendlight}.
To obtain the signal control policy, we first transform the state vector $S_t$ for each agent (i.e., each intersection) at decision time step $t$ into a high-dimensional embedding vector through a Multi-layer Perceptron (MLP), which comprises two linear layers. The resulting state embedding vector is represented as:
\begin{equation}
{h}_{s} \in \mathbb{R}^{d} =\mathbf{MLP}^{(2)}_{s} \left(S_t\right),
\end{equation}
\noindent where $d$ denotes the embedding dimension. 
The embedding vector is subsequently fed into a recurrent neural network, specifically a Gated Recurrent Unit (GRU)~\cite{chung2014empirical}, which selectively incorporates historical traffic state information while ignoring the irrelevant ones. 
The aggregated state embedding vector $\hat{{h}}_{s}$ is derived as follows:
\begin{equation}
\hat{{h}}_{s} \in \mathbb{R}^{d} =\mathbf{GRU} \left({h}_{s}, {h}^{GRU}_{(s, t-1)} \right),
\end{equation}
\noindent where ${h}^{GRU}_{t-1}$ denotes the hidden vector produced by the GRU at the previous decision time step $t-1$. 
Similarly, we transform the comprehensive phase state vector $G$ for the intersection into a high-dimensional embedding vector using an additional MLP composed of two linear layers. The resulting phase embedding vector ${h}_{p}$ is then calculated as:
\begin{equation}
{h}_{p} \in \mathbb{R}^{|\mathcal{P}| \times d} =\mathbf{MLP}^{(2)}_{p}\left({G} \right).
\end{equation}
We further employ a multi-head cross-attention mechanism adopted from Transformer architecture~\cite{vaswani2017attention} to generate phase-conditioned state embedding vectors. 
Specifically, for each head $h$ (in total 4 heads), the query vector ${Q}_{h}$ is derived from the phase embedding vector ${h}_{p}$, the key vector ${K}_{h}$ and the value vector ${V}_{h}$ are generated from the aggregated state embedding vector $\hat{{h}}_{s}$ produced by the GRU. The query, key, and value vectors are calculated as:
$\textbf{Query}: {Q}_{h} \in \mathbb{R}^{|\mathcal{P}| \times d} = {h}_{p} \, W_h^Q$, $\textbf{Key}: {K}_{h} \in \mathbb{R}^{1 \times d} = \hat{{h}}_{s} \, W_h^K$, and  $\textbf{Value}: {V}_{h} \in \mathbb{R}^{1 \times d} = \hat{{h}}_{s} \, W_h^V$.
Here, $W_h^Q$, $W_h^K$, and $W_h^V$ are the learnable parameters of the attention mechanism of head $h$.
The attention vector for each head is computed as:
\begin{equation}
\operatorname{Att}_h({Q_h}, {K_h},{V_h})=\operatorname{softmax}\left(\frac{{Q}_h\left({K}_h\right)^T}{\sqrt{d}}\right) {V}_h.
\end{equation}
The advanced state embedding vector ${h}_{sp}$ is obtained by concatenating the attention vectors from all heads and passing them through a linear layer with parameters $ W^O$: ${h}_{sp} \in \mathbb{R}^{|\mathcal{P}| \times d} = \operatorname{Concat}\left(\text{Att}_1, \text{Att}_2, \text{Att}_3, \text{Att}_4\right) \; W^O$.

\begin{figure*}[t!]
    \centering
    \includegraphics[
    width=\linewidth]{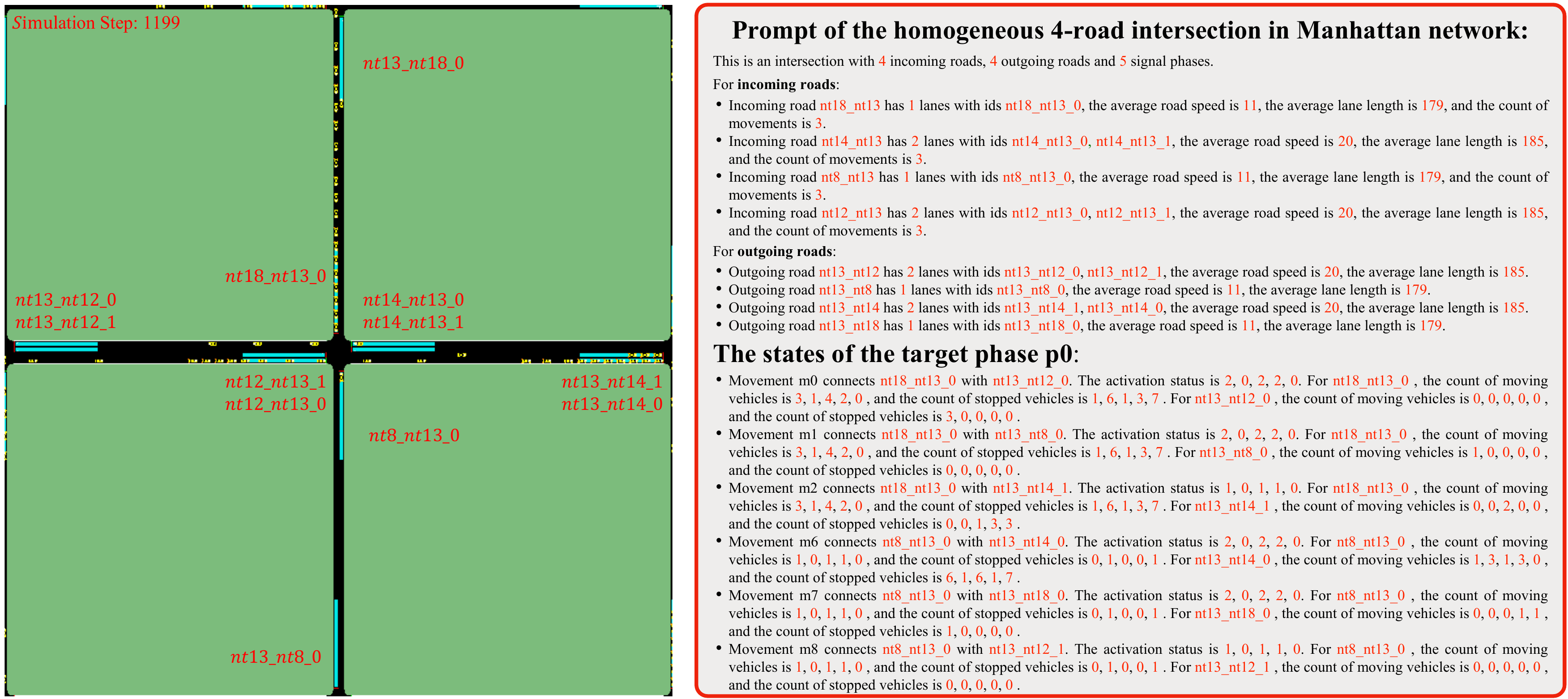}
    \caption{
    A snapshot of a four-road intersection from the Grid 5$\times$5 network during the simulation for generating semantic features in LATS (left).
    The traffic prompt for phase $p0$ at the intersection, which contains the descriptions of static intersection topology and dynamic traffic conditions (right). The black text represents the unified prompt template, while the orange text indicates the filled-in intersection topology and traffic dynamics information.}
    \label{fig:phase_prompt}
\end{figure*}

\subsection{LLM-assisted Teacher-Student Module}
The complexity of dynamic traffic demands and diverse intersection topologies poses significant challenges in learning efficient general control strategies due to limited model representation capability.
Embedding LLMs, with their exceptional understanding and generalization abilities, can provide high-quality semantic representations at intersections.
By using textual descriptions of both static topology information and dynamic traffic conditions, embedding LLMs provides unique semantic insights and enhances phase representations for RL decision-making. 
This integration of multi-modal features greatly improves the model's representation capability, leading to more efficient and generalizable shared strategies. 
Additionally, we propose a plug-and-play teacher-student module, as illustrated in Fig.~\ref{fig:method}, which transfers the LLM’s semantic representation capability to a simpler student network through latent space alignment.
We also demonstrate its flexibility by integrating it with other general TSC methods, such as AttendLight~\cite{oroojlooy2020attendlight}, to enhance their control performance, as discussed in the ablation study (refer to Section~\ref{sec:ablation_study}).

In this teacher-student module, we combine the time-variant state vector $S_t$, time-invariant phase vector $G_p$, and time-invariant topology vector $I$ to form a vector for each phase $p$ at the intersection: $x_{p}(t)=[S_t, G_p, I]$. 
This vector is then used to automatically produce a customized traffic prompt by filling in a pre-defined template. 
The prompt design follows two key guidelines, completeness and conciseness, which ensure that each prompt provides sufficient topological and traffic information while remaining concise and focused to preserve the distinctiveness of the generated embeddings.

As shown on the left side of Fig.~\ref{fig:method}, the template shared across all intersections consists of two parts. 
The first part depicts the intersection's topology, including the type of intersection, the count of available phases, and the details of all roads and lanes. 
The second part describes the states of the involved traffic movements given the signal phase $p$, which includes the movement activation status and the counts of moving and stopped vehicles on the incoming and connected outgoing lanes. 
To better capture traffic dynamics, we include traffic states from the previous four steps to create a temporal state prompt, since this choice offers a practical balance between capturing meaningful temporal context and avoiding redundant information, as further supported by the ablation study in the \href{https://www.dropbox.com/scl/fi/zf1c7p81ua8cc6x6vzssm/Supplementary_Materials_T-ITS_LATS.pdf?rlkey=zfa2elenxz383ien27h63alja&st=adtp54jm&dl=0}{supplementary materials}.
This prompt describes both the static and dynamic semantic characteristics of the intersection to provide rich contextual information.
An illustration of the prompt for phase $p0$ of the four-road intersection in the Grid 5$\times$5 network is detailed in Fig.~\ref{fig:phase_prompt}. 
For each phase at the intersection, we use a unified template (shown in black text) to generate the corresponding phase prompt, which shares the same intersection topology description but includes a unique traffic dynamics description for the involved traffic movements (shown in orange text).
Additionally, we present the prompts for a more realistic three-road intersection with two phases and a four-road intersection with four phases in the heterogeneous Monaco network in the \href{https://www.dropbox.com/scl/fi/zf1c7p81ua8cc6x6vzssm/Supplementary_Materials_T-ITS_LATS.pdf?rlkey=zfa2elenxz383ien27h63alja&st=adtp54jm&dl=0}{supplementary materials}.

Such prompts can be further enriched by adding more contextual information, such as weather, time of day, special events, and historical traffic patterns, to generate richer semantic representations via LLMs.
We then input the resulting text prompt to a pre-trained embedding $\mathbf{LLM}^{*}$, specifically \textit{jina-embeddings-v2-small-en}\footnote{\url{https://huggingface.co/jinaai/jina-embeddings-v2-small-en}}~\cite{gunther2023jina}.
This open-source long-context embedding model, containing approximately 33 million parameters, is built on a modified BERT architecture with ALiBi attention and trained through large-scale pre-training and multi-stage fine-tuning on diverse text corpora to capture general-purpose semantic representations.
It generates a semantic embedding vector ${e}_p \in \mathbb{R}^{d'}$, represented as:
\begin{equation}
{e}_p = \mathbf{LLM}^{*}(Prompt(x_p(t))).
\end{equation}
Following the same sequence in constructing the phase state vector $G$, we obtain the semantic embedding vectors for all available phases at the intersection as: ${e} \in \mathbb{R}^{|\mathcal{P}| \times d^{'}}=[{e}_p \mid p \in \mathcal{P}]$,
where $d^{'}=512$ is the embedding dimension of $\mathbf{LLM}^{*}$. 
The embedding LLM, serving as the teacher model, guides the student model by providing rich semantic features, which the student model learns to emulate through knowledge distillation at the latent space, ensuring more effective feature alignment and learning efficiency.
To this end, we apply variational inference techniques, specifically two Variational AutoEncoders (VAEs)~\cite{kingma2013auto}, to obtain a 32-dimensional latent variable $z_p$ of phase $p$, which is derived from a parameterized distribution with a mean vector $z_p(\mu)$ and variance vector $z_p(\sigma)$, for both teacher model and student model. The detailed architecture of the VAE is illustrated in Fig.~\ref{fig:method}, consisting of the following components with the step $t$ omitted for simplicity,
\begin{equation}
\begin{aligned}
\textit{Encoder}: &\quad q\left(z_p^s \mid x_p ; \psi_e\right), q\left(z_p^c \mid e_p ; \omega_e\right),\\
\textit{Decoder}: &\quad p\left(\hat{x}_p \mid z_p^s ; \psi_d\right), p\left(\hat{e}_p \mid z_p^c ; \omega_d\right).\\
\end{aligned}
\end{equation}
Here, the student's VAE encoder and decoder are parameterized by $\psi_e$ and $\psi_d$, respectively, while the teacher's VAE encoder and decoder are parameterized by $\omega_e$ and $\omega_d$.
The decoders output $\hat{x}_p$ and $\hat{e}_p$ to reconstruct the inputs $x_p$ and $e_p$.
The Evidence Lower Bound (ELBO) is the objective function that a VAE maximizes to balance reconstruction accuracy and latent space regularization, which is expressed as:
\begin{equation}
\begin{aligned}
    \operatorname{ELBO}(\psi_e, \psi_d)&=\mathbb{E}_{q(z_p^s \mid x_p ; \psi_e)}[\log p(x_p \mid z_p^s ; \psi_d)]\\
    &-\mathrm{KL}(q(z_p^s \mid x_p ; \psi_e) \;\|\; p(z_p^s)),\\
    \operatorname{ELBO}(\omega_e, \omega_d)&=\mathbb{E}_{q(z_p^c \mid e_p ; \omega_e)}[\log p(e_p \mid z_p^c ; \omega_d)]\\
    &-\mathrm{KL}(q(z_p^c \mid e_p ; \omega_e) \;\|\; p(z_p^c)).
\end{aligned}
\end{equation}
Here, the prior distribution of latent variables of teacher model and student model are assumed to be the standard normal distribution: $p\left(z_p^s\right) = \textit{N}(z_p^s; \textbf{0}, \textbf{I})$, and $p\left(z_p^c\right)=\textit{N}(z_p^c; \textbf{0}, \textbf{I})$. 
Furthermore, we align the distributions of latent variables derived from the student model with the distributions of latent variables derived from the teacher model to achieve knowledge distillation and impartation. This is achieved by optimizing the parameters of the student model to minimize the KL divergence between the distributions of the latent variables derived from teacher and student models: $\mathrm{KL}(z_p^s, z_p^c)=\mathrm{KL}(\mathcal{N}(z_p^c(\mu), z_p^c(\sigma))\;\|\;\mathcal{N}(z_p^s(\mu), z_p^s(\sigma)))$.
Thus, the final optimization objective for the teacher-student learning module in LATS to maximize is defined as:
\begin{equation}
\begin{aligned}
\label{elbo_ts}
    J(\psi, \omega)=\underbrace{ELBO(\psi_e, \psi_d, \omega_e, \omega_d)}_{\textit{Sum of Teacher-Student ELBO}} - \underbrace{\mathrm{KL}(z_p^s(\psi_e), z_p^c(\omega_e))}_{\textit{Latent Alignment Loss}}.
\end{aligned}
\end{equation}

The computational complexity of our teacher–student module is mainly dominated by repeated LLM embedding queries, whereas the VAEs in both the teacher and student networks introduce only minor additional overhead. 
A detailed analysis and discussion are provided in the \href{https://www.dropbox.com/scl/fi/zf1c7p81ua8cc6x6vzssm/Supplementary_Materials_T-ITS_LATS.pdf?rlkey=zfa2elenxz383ien27h63alja&st=adtp54jm&dl=0}{supplementary materials}.

\begin{table*}
\caption{Comparative results of different methods on both Grid 5$\times$5 and Monaco networks across average episodic metrics: queue length ($\downarrow$), speed ($\uparrow$), intersection delay ($\downarrow$), completion rate ($\uparrow$), trip time ($\downarrow$), and trip delay ($\downarrow$). Here, $\uparrow$ indicates higher values are better, and $\downarrow$ indicates lower values are better, with the best result for each metric in bold and the second best underlined.}
\label{exp:exp1}
\resizebox{\textwidth}{!}{%
\begin{tabular}{c|cccccc}
\hline 
& \begin{tabular}[c]{@{}c@{}}Queue Length $\downarrow$\\ (vehs)\end{tabular} &
  \begin{tabular}[c]{@{}c@{}}Speed $\uparrow$ \\(m/s)\end{tabular} &
  \begin{tabular}[c]{@{}c@{}}Intersections Delay $\downarrow$\\ (sec)\end{tabular} &
  \begin{tabular}[c]{@{}c@{}}Completion Rate $\uparrow$ \\(vehs/sec)\end{tabular} &
  \begin{tabular}[c]{@{}c@{}}Trip Time $\downarrow$ \\ (sec)\end{tabular} &
  \begin{tabular}[c]{@{}c@{}}Trip Delay $\downarrow$ \\ (sec)\end{tabular} \\ 
\hline
\multicolumn{7}{c}{Homogeneous Grid 5$\times$5 network with 25 signalized intersections} \\ \hline
Fixed-Time &
  3.16 (1.57) &
  2.03 (1.56) &
  29.34 (17.65) &
  0.62 (0.54) &
  657.67 (551.72) &
  411.95 (428.68) \\
Greedy &
  2.65 (1.72) &
  3.27 (2.42) &
  42.09 (38.05) &
  0.86 (0.41) &
  546.96 (429.35) &
  263.59 (325.64) \\
Max-Pressure &
  2.79 (1.76) &
  2.93 (2.19) &
  39.17 (37.50) &
  0.82 (0.42) &
  572.43 (404.92) &
  270.41 (292.18) \\ \hline
IQLD &
  3.51 (2.24) &
  2.36 (2.12) &
  95.92 (74.70) &
  0.58 (0.39) &
  492.71 (439.87) &
  285.46 (368.77) \\
IQLL &
  2.00 (1.23) &
  3.38 (2.32) &
  40.60 (40.68) &
  0.95 (0.47) &
  483.48 (406.18) &
  240.27 (300.38) \\
IA2C &
  3.94 (2.29) &
  1.52 (1.09) &
  81.49 (48.96) &
  0.40 (0.26) &
  760.47 (510.71) &
  546.28 (446.36) \\
MA2C &
   2.83 (1.49) &
   2.14 (1.06) &
   38.44 (23.92) &
   0.65 (0.38) &
   595.30 (432.10) &
   383.86 (346.76) \\ \hline
Llama-2-7b &
   4.54 (2.53) &
   1.95 (1.52) &
   48.37 (27.81) &
   0.63 (0.80) &
   790.17 (491.24) &
   539.55 (416.21) \\ 
Llama-3-8B &
   4.06 (2.29) &
   2.07 (1.67) &
   66.72 (44.43) &
   0.64 (0.79) &
   882.89 (592.03) &
   639.23 (526.49) \\ \hline
IPPO &
   4.30 (2.44) &
   1.92 (1.44) &
   53.36 (34.60) &
   0.60 (0.33) &
   935.52 (621.47) &
   687.05 (563.01) \\
AttendLight &
   3.37 (2.17) &
   3.22 (2.55) &
   62.89 (46.97) &
   0.85 (0.37) &
   569.80 (491.76) &
   358.32 (426.50) \\
HolLight$^{*}$ &
    2.47 (1.58) &
    3.40 (2.40) &
    \underline{18.77 (12.62)} &
    0.94 (0.44) &
    607.59 (453.83) &
    312.24 (335.18) \\
CoLight$^{*}$&
    1.63 (1.14) &
    4.52 (2.26) &
    22.39 (15.63) &
    \underline{1.01 (0.50)} &
    \underline{425.31 (347.22)} &
    \underline{216.29 (279.09)} \\
HeteroLight &
   \underline{1.57 (1.18)} &
   \underline{4.55 (2.29)} &
   22.18 (16.94) &
   \textbf{1.03 (0.49)} &
   443.30 (351.48) &
   228.67 (283.31) \\
LATS (ours) &
  \textbf{1.26 (1.04)} &
  \textbf{4.73 (2.39)} &
  \textbf{13.60 (9.91)} &
  \textbf{1.03 (0.58)} &
  \textbf{389.80 (288.08)} &
  \textbf{182.72 (218.25)} \\ \hline
\multicolumn{7}{c}{Heterogeneous Monaco network with 28 signalized intersections} \\ \hline
Fixed-Time &
    2.09 (1.25) &
    1.69 (1.83) &
    \textbf{66.41 (53.36)} &
    0.26 (0.24) &
    652.17 (531.78) &
    465.60 (484.56) \\
Greedy &
    2.08 (1.40) &
    2.41 (3.04) &
    87.82 (63.38) &
    0.20 (0.21) &
    529.68 (615.31) &
    377.37 (577.15) \\
Max-Pressure &
    2.01 (1.33) &
    2.42 (3.01) &
    85.71 (61.90) &
    0.21 (0.21) &
    545.64 (608.07) &
    392.80 (571.18) \\ \hline
IQLD &
    2.02 (1.06) &
    0.79 (1.10) &
    116.92 (34.36) &
    0.11 (0.14) &
    587.76 (664.27) &
    464.95 (640.21) \\
IQLL &
    1.27 (0.83) &
    2.49 (2.69) &
    86.82 (50.28) &
    0.28 (0.22) &
    419.45 (508.70) &
    293.68 (484.05) \\
IA2C &
    \underline{1.08 (0.56)} &
    2.07 (2.14) &
    82.94 (35.56) &
    0.27 (0.18) &
    410.69 (507.16) &
    296.70 (486.29) \\
MA2C &
     1.13 (0.74) &
     3.20 (3.64) &
     79.60 (57.01)&
     0.31 (0.23) &
     384.85 (490.66) &
     261.57 (467.18) \\ \hline
Llama-2-7b &
    2.06 (1.05) &
    1.16 (1.54) &
    83.22 (40.06) &
    0.17 (0.41) &
    778.38 (629.17) &
    630.53 (587.79) \\
Llama-3-8B &
    1.75 (0.98) &
    2.18 (2.86) &
    74.72 (48.64) &
    0.26 (0.50) &
    722.57 (628.14) &
    568.12 (585.41) \\ \hline
IPPO &
    1.40 (0.78) &
    2.37 (2.29) &
    73.70 (50.90) &
    0.34 (0.18) &
    777.14 (678.98) &
    627.99 (670.14) \\
AttendLight &
    1.46 (1.05) &
    3.38 (3.58) &
    \underline{69.41 (56.94)} &
    \underline{0.37 (0.25)} &
    484.66 (503.37) &
    346.26 (475.15) \\
HolLight$^{*}$ &
    1.45 (1.03) &
    \underline{3.63 (4.20)} &
    70.78 (55.65) &
    0.31 (0.24) &
    440.43 (515.72) &
    306.15 (486.24) \\
CoLight$^{*}$ &
    1.28 (0.82) &
    3.62 (4.27) &
    78.29 (51.77) &
    0.32 (0.22) &
    413.19 (548.33) &
    287.63 (519.49) \\
HeteroLight &
    1.22 (0.76) &
    3.45 (3.84) &
    102.56 (73.40) &
    0.34 (0.23) &
    \underline{383.87 (483.26)} &
    \underline{252.11 (459.32)} \\
LATS (ours) &
    \textbf{1.05 (0.66)} &
    \textbf{3.87 (4.05)} &
    76.48 (52.33) &
    \textbf{0.39 (0.24)} &
    \textbf{358.41 (431.60)} &
    \textbf{230.38 (408.35)} \\ \hline
\end{tabular}}
\end{table*}

\subsection{Policy Optimization}
We then integrate the learned semantic features into the RL decision-making process. 
We input the latent mean vectors from the student model into a two-layer MLP to calculate the semantic embedding vector of all available phases used for decision-making at the intersection, which is written as:
\begin{equation}
    e_{sp} \in \mathbb{R}^{|\mathcal{P}| \times d} = \textbf{MLP}_{sp}^{(2)}([z_p^s(\mu) \mid p \in \mathcal{P}]).
\end{equation}
We concatenate the semantic embedding vector $\mathbf{e}_{sp}$ with the advanced state embedding vector $\mathbf{h}_{sp}$ to form $\widetilde{h}_{sp}=[{h}_{sp}||{e}_{sp}]$. 
The policy and value functions are then computed by passing $\widetilde{h}_{sp} \in \mathbb{R}^{|\mathcal{P}| \times 2d}$ through two different one-layer MLPs, each mapping from $2d$ to $1$ dimension, calculated as:
\begin{equation}
\begin{aligned}
        \mathbf{\pi} \in \mathbb{R}^{|\mathcal{P}| \times 1} &= \operatorname{Softmax}(\mathbf{MLP}_{\pi}^{(1)}(\widetilde{h}_{sp})),\\
        {V} \in \mathbb{R}^{1} &= \operatorname{Sum}(\mathbf{MLP}_{v}^{(1)}(\widetilde{h}_{sp})).
\end{aligned}
\end{equation}
Additionally, we generate a vector $\hat{S}$ from a separate one-layer MLP to predict $S_{t+1}$, the state at the next time step using the combined state and semantic embedding vector $\widetilde{h}_{sp}$:
\begin{equation}
    \hat{S} = \mathbf{MLP}_{pred}^{(1)}(\widetilde{h}_{sp}).
\end{equation}
We pad the state vector and phase vector of different intersections to the same dimension to facilitate batch training.
Additionally, we apply a mask to filter out those unavailable actions during the phase selection process. We use the Proximal Policy Optimization (PPO)~\cite{schulman2017proximal} algorithm to update the policy and value functions by calculating the policy loss $L_{\theta}$, value loss $L_{\phi}$, entropy loss $L^{e}_{\theta}$, and prediction loss $L^{au}_{\theta, \phi}$.
For each agent $i$, the loss function $L_{\theta, \tau_i}$ for optimizing the policy function parameterized by $\theta$ is written as follows:
\begin{equation}
-{\mathbb{E}}_{\tau_i}\left[\min \left(r_{\tau_i}(\theta) \hat{A}_{\tau_i}, \operatorname{clip}\left(r_{\tau_i}(\theta), 1-\epsilon, 1+\epsilon\right) \hat{A}_{\tau_i}\right)\right],
\end{equation}
where $r_{\tau_i}(\theta)=\frac{\pi_\theta\left(a_{\tau_i} \mid \cdot \right)}{\pi_{\theta_{\text {old}}}\left(a_{\tau_i} \mid \cdot \right)}$ is the probability ratio of the new policy to the old policy given trajectory $\tau_i$. 
While $\hat{A}_{\tau_i}$ is the advantage estimate, computed using the General Advantage Estimate (GAE)~\cite{schulman2015high}. 
To encourage exploration during training, we add an entropy bonus, calculated as $L_\theta^e={\mathbb{E}}_{\tau_i}\left[\mathcal{H}\left(\pi_\theta\left(\cdot\right)\right)\right]$, where $\mathcal{H}\left(\pi_\theta\left(\cdot\right)\right)$ is the policy entropy.
This encourages the policy to stay stochastic, avoid premature convergence, and better explore the state-action space.
Additionally, let $V_{\phi}$ and $V^{'}_{\phi}$ denote the state values of the current and next steps, respectively. The loss function for the value function is calculated based on the temporal difference (TD) error:
\begin{equation}
\label{eqn:value_loss}
    L_{\phi, \tau_i}={\mathbb{E}}_{\tau_i}\left[\left(\underbrace{R_{\tau_i} + \gamma \, V^{'}_{\phi, \tau_i}}_{\textit{target values}} - V_{\phi, \tau_i} \right)^2\right],
\end{equation}
where $R$ is the individual reward. To further enhance the model's understanding of traffic dynamics transitions, we define a prediction loss function based on the Mean Square Error (MSE) between the model's predicted state vector $\hat{S}$, obtained from the prediction head, and the actual next traffic state $S'$. The prediction loss is computed as:
\begin{equation}
L^{au}_{\theta, \phi, \tau_i}={\mathbb{E}}_{\tau_i}\left[\left(\hat{S}_{\tau_i} - S^{'}_{\tau_i} \right)^2\right].
\end{equation}
Thus, the overall RL loss function can be computed as:
\begin{equation}
\label{eqn:rl_loss}
L^{rl}_{\theta, \phi}(\tau_i)= L_{\theta, \tau_i} + c_1 L_{\phi, \tau_i} - c_2 L^{e}_{\theta, \tau_i} + c_3 L^{au}_{\theta, \phi, \tau_i},
\end{equation}
where $c_1$, $c_2$, and $c_3$ are the weighting coefficients for the value loss, entropy loss, and prediction loss, respectively.  By combining the RL losses with the teacher-student module loss $L^{ts}_{\psi, \omega}=-J(\psi, \omega)$ derived from Eq(\ref{elbo_ts}), the total optimization loss function for all $N$ agents is expressed as:
\begin{equation}
\label{eqn:total_loss}
L(\theta, \phi, \psi, \omega;\tau)=\frac{1}{N}\sum_{i=1}^N\left(L^{rl}_{\theta, \phi}(\tau_i) + L^{ts}_{\psi, \omega}(\tau_i) \right).
\end{equation}
The pseudocode of LATS' optimization algorithm, including the LLM-assisted teacher-student learning and the overall policy optimization process, are provided in the \href{https://www.dropbox.com/scl/fi/zf1c7p81ua8cc6x6vzssm/Supplementary_Materials_T-ITS_LATS.pdf?rlkey=zfa2elenxz383ien27h63alja&st=adtp54jm&dl=0}{supplementary materials}.
Notably, optimizing the LLM-assisted teacher-student module and the RL policy can be executed concurrently, using the same set of collected trajectories for training.

\begin{table*}[t]
\centering
\caption{Zero-shot performance of various parameter sharing methods trained on the Monaco network and tested on the \textit{Grid 5$\times$5} network with low, medium, and high demands.}
\label{exp:exp2}
\resizebox{\textwidth}{!}{%
\begin{tabular}{c|cccccc}
\hline &
  \begin{tabular}[c]{@{}c@{}}Queue Length $\downarrow$ \\ (vehs)\end{tabular} &
  \begin{tabular}[c]{@{}c@{}}Speed $\uparrow$ \\ (m/s)\end{tabular} &
  \begin{tabular}[c]{@{}c@{}}Intersection Delay $\downarrow$ \\ (sec)\end{tabular} &
  \begin{tabular}[c]{@{}c@{}}Completion Rate $\uparrow$ \\ (vehs/s)\end{tabular} &
  \begin{tabular}[c]{@{}c@{}}Trip Time $\downarrow$ \\ (sec)\end{tabular} &
  \begin{tabular}[c]{@{}c@{}}Trip Delay $\downarrow$ \\ (sec)\end{tabular} \\ \hline
\multicolumn{7}{c}{Low traffic demands for \textit{Grid 5$\times$5} network} \\ \hline
IPPO & 0.68 (0.36) & 2.46 (1.03) & 61.95 (20.47) & 0.20 (0.18) & 625.72 (499.98) & 472.59 (475.34) \\
AttendLight & \underline{0.23 (0.12)} & \underline{4.17 (1.74)} & \underline{42.23 (23.15)} & \underline{0.21 (0.21)} & \underline{299.15 (184.76)} & \underline{160.65 (180.29)} \\
HeteroLight & 0.47 (0.21) & 3.19 (1.98) & 227.85 (160.84) & 0.20 (0.20) & 441.39 (705.13) & 326.66 (707.81) \\
LATS (ours) & \textbf{0.12 (0.12)} & \textbf{6.34 (2.94)} & \textbf{14.10 (10.12)} & \textbf{0.22 (0.22)} & \textbf{209.64 (167.81)} & \textbf{83.65 (148.05)} \\ \hline
\multicolumn{7}{c}{Medium traffic demands for \textit{Grid 5$\times$5} network} \\ \hline
IPPO & 2.58 (1.19) & 1.68 (1.05) & 131.46 (67.17) & 0.37 (0.25) & 882.97 (754.10) & 717.94 (727.24) \\
AttendLight & \underline{1.72 (0.94)} & \underline{2.85 (1.41)} & \textbf{56.22 (25.75)} & \underline{0.57 (0.35)} & \underline{614.12 (513.10)} & \underline{415.40 (474.18)} \\
HeteroLight & 2.13 (1.22) & 2.56 (2.09) & 217.84 (139.27) & 0.43 (0.31) & 690.88 (823.80) & 556.44 (832.08) \\
LATS (ours) & \textbf{1.63 (0.86)} & \textbf{2.86 (1.40)} & \underline{79.95 (38.96)} & \textbf{0.58 (0.39)} & \textbf{515.51 (464.00)} & \textbf{335.81 (414.19)} \\ \hline
\multicolumn{7}{c}{High traffic demands for \textit{Grid 5$\times$5} network} \\ \hline
IPPO & \textbf{3.78 (1.80)} & 1.48 (1.03) & 138.42 (68.45) & 0.44 (0.27) & 1010.58 (873.64) & 834.15 (840.43) \\
AttendLight & 4.08 (2.15) & \textbf{2.04 (1.70)} & \textbf{84.56 (40.37)} & \textbf{0.57 (0.30)} & \underline{910.05 (758.20)} & \underline{714.63 (719.66)} \\
HeteroLight & 5.21 (3.13) & 1.82 (2.05) & 195.25 (119.39) & 0.43 (0.26) & 1145.30 (930.94) & 958.58 (927.32) \\
LATS (ours) & \underline{3.79 (2.05)} & \underline{1.92 (1.42)} & \underline{106.07 (41.37)} & \underline{0.56 (0.30)} & \textbf{681.23 (612.32)} & \textbf{482.87 (560.74)} \\ \hline
\end{tabular}}
\end{table*}

\begin{table*}[t!]
\centering
\caption{
Zero-shot performance of various parameter sharing RL methods trained on the homogeneous \textit{Grid 5$\times$5} network and tested on the heterogeneous Monaco network.
}
\label{exp:c1_exp2_grid_to_monaco}
\resizebox{\textwidth}{!}{%
\begin{tabular}{c|cccccc}
\hline &
  \begin{tabular}[c]{@{}c@{}}Queue Length $\downarrow$ \\ (vehs)\end{tabular} &
  \begin{tabular}[c]{@{}c@{}}Speed $\uparrow$ \\ (m/s)\end{tabular} &
  \begin{tabular}[c]{@{}c@{}}Intersection Delay $\downarrow$ \\ (sec)\end{tabular} &
  \begin{tabular}[c]{@{}c@{}}Completion Rate $\uparrow$ \\ (vehs/s)\end{tabular} &
  \begin{tabular}[c]{@{}c@{}}Trip Time $\downarrow$ \\ (sec)\end{tabular} &
  \begin{tabular}[c]{@{}c@{}}Trip Delay $\downarrow$ \\ (sec)\end{tabular} \\ \hline
AttendLight  & 2.14 (1.28) & \underline{1.67 (2.36)} & \textbf{85.29 (43.15)} & 0.15 (0.14) & 1372.88 (875.09) & 1242.56 (850.96) \\
HeteroLight  & \underline{1.91 (1.10)} & \textbf{1.78 (2.39)} & \underline{88.55 (43.04)} & \underline{0.17 (0.14)} & \underline{1272.23 (874.33)} & \underline{1145.17 (855.98)} \\
LATS (ours)  & \textbf{1.60 (0.90)} & 1.58 (2.04) & 99.18 (36.52) & \textbf{0.19 (0.15)} & \textbf{478.01 (561.71)} & \textbf{351.49 (535.29)} \\ \hline
\end{tabular}}
\end{table*}

\section{EXPERIMENTS AND RESULTS}
In this section, we conduct experiments and evaluate the effectiveness of our proposed method, LATS, by answering the following three key questions: 
(1) \textbf{Q1}: How does the proposed LLM-assisted teacher-student framework perform compared to existing ATSC methods in optimizing traffic flow?
(2) \textbf{Q2}: What advantages does our teacher-student approach provide for enhancing generalizability andtransferability across diverse traffic scenarios?
(3) \textbf{Q3}: How does the integration of LLMs enhance decision-making in TSC tasks?

\subsection{Experimental Setup}
We conduct simulation experiments using the open-source microscopic traffic simulator, Simulation of Urban MObility (SUMO)~\cite{SUMO2018}, utilizing traffic datasets adopted from the prior MA2C work~\cite{chu2019multi}. 
The dataset comprises two traffic networks along with their corresponding traffic demands. The first network, Grid 5$\times$5, is a homogeneous grid with 25 homogeneous intersections. The second network, Monaco, is a real-world network featuring 28 heterogeneous intersections.
Detailed specifics of the intersections within the Grid 5$\times$5 and Monaco networks, as well as the synthetic traffic demands including start times and Origin-Destination (OD) pairs for various dense traffic flows, are provided in the \href{https://www.dropbox.com/scl/fi/zf1c7p81ua8cc6x6vzssm/Supplementary_Materials_T-ITS_LATS.pdf?rlkey=zfa2elenxz383ien27h63alja&st=adtp54jm&dl=0}{supplementary materials}.

\subsection{Baseline Methods and Evaluation Metrics}
We select the following TSC baselines to ensure a comprehensive comparison of performance across different methods:
\begin{itemize}
    \item \textbf{Traditional TSC methods}: Fixed-time~\cite{roess2004traffic}, Greedy~\cite{goel2023sociallight}, and Max-pressure~\cite{kouvelas2014maximum}.
    \item \textbf{Pure-LLM-based TSC methods}: LLMLight~\cite{lai2023large} that directly uses conversational LLMs (e.g., Llama-2-7b and Llama-3-8B~\cite{touvron2023llama}) for decision-making.
    \item \textbf{Pure-RL-based TSC methods}: Independent learning methods, including MA2C and its baselines IQLL, IQLD, and IA2C, as well as parameter-sharing RL methods, such as IPPO~\cite{schulman2017proximal}, CoLight$^{*}$~\cite{wei2019colight}, AttendLight~\cite{oroojlooy2020attendlight}, HolLight$^{*}$~\cite{qiao2024hol}, and HeteroLight~\cite{zhang2024HeteroLight}.
    Methods marked with $^{*}$ are reimplemented using the PPO algorithm for a fair comparison within the same RL framework.
    \item \textbf{Variants for ablation}: LATS w/o T (without the teacher network), LATS w/o S (without the student network), LATS w/o TS (without the entire teacher-student module), and AttendLight w/ TS (AttendLight integrated with the teacher-student module) for the component ablation study; LATS w/ Topo (with the topology-only prompt) and LATS w/ Dyna (with the traffic-dynamics-only prompt) for the prompt ablation study; LATS w/ BGE (replacing the \textit{Jina} embedding LLM with the \textit{BGE} model~\cite{bge_embedding}) for the embedding LLM ablation study.
\end{itemize}

In all experiments, we adopt a 10-second green phase and a 3-second yellow phase. 
If the same phase is maintained, the signal remains green for 10 seconds; if the phase changes, it transitions to a 3-second yellow followed by a 7-second green, resulting in the same 10-second duration. 
Each episode lasts for 3600 seconds (360 decision steps), following the MA2C~\cite{chu2019multi} setting. 
All baselines are trained with their official implementations or under the same configuration as LATS, and are run until convergence before evaluation.
The experiments are conducted on an Ubuntu server with 32GB RAM, an Intel Core i9-13900KF processor (24 cores, 32 threads), and an NVIDIA RTX 4090 GPU.
The full LATS training takes around 24 hours per 1000 episodes, mainly due to the embedding LLM inference, while other RL baselines such as HeteroLight and AttendLight require about 4 hours under the same setup.
We then evaluate all methods over 10 episodes using different random seeds (consistent across methods) and report the evaluation performance as the mean and standard deviation, following the testing protocol of MA2C~\cite{chu2019multi}. 
The evaluation metrics include average queue length, average speed, average intersection delay, completion rate, average trip time, and average trip delay. 
Details of all baselines, the training parameters for all RL-based methods, and the evaluation metrics are provided in the \href{https://www.dropbox.com/scl/fi/zf1c7p81ua8cc6x6vzssm/Supplementary_Materials_T-ITS_LATS.pdf?rlkey=zfa2elenxz383ien27h63alja&st=adtp54jm&dl=0}{supplementary materials}.

\begin{table}[t!]
\caption{
Comparison of Average Trip Duration (s) across different ATSC methods on the homogeneous Grid 5$\times$5 network and the heterogeneous Monaco network. 
}
\label{exp:exp5}
\resizebox{0.45\textwidth}{!}{%
\begin{tabular}{c|cc}
\hline
& \begin{tabular}[c]{@{}c@{}} Grid $5\times5$ Network\\\end{tabular} &
  \begin{tabular}[c]{@{}c@{}} Monaco Network\\ \end{tabular} \\
\hline
Fixed-Time    & 1145.72 (362.93) & 1117.75 (357.76) \\
Greedy        & 679.29 (235.53) & 1025.35 (325.50) \\
Max-Pressure  & 914.37 (312.51) & 1027.53 (329.69) \\ \hline
AttendLight   & 767.36 (255.02) & 909.55 (288.93) \\
HolLight*     & 622.11 (215.92) & 920.95 (297.80) \\
CoLight*      & \underline{464.27 (161.76)} & 884.26 (282.46) \\
HeteroLight   & 477.15 (173.31) & \underline{874.64 (280.68)} \\
LATS (ours)          & \textbf{424.77 (139.91)} & \textbf{771.71 (250.81)} \\
\hline
\end{tabular}}
\centering
\vspace{-0.5cm}
\end{table}

\subsection{Overall Performance (Q1)}
Considering the results for the challenging homogeneous Grid 5$\times$5 network, as shown in Table~\ref{exp:exp1}, our method LATS demonstrates superior performance across six traffic metrics. 
LATS achieves the lowest average queue length, significantly outperforming traditional methods (Fixed-Time, Greedy, and Max-Pressure), independent learning methods (IQLD, IQLL, IA2C, and MA2C), pure-LLM-based methods, and parameter-sharing methods (IPPO, AttendLight, HolLight, CoLight, and HeteroLight). 
It also achieves the highest average speed and the lowest intersection delay, indicating its effectiveness in optimizing traffic flow.
Regarding completion rate, trip time, and trip delay, LATS demonstrates the lowest trip time (389.80 seconds) and trip delay (182.72 seconds), along with the highest completion rate (1.03 veh/s). These metrics are significantly better than those of traditional methods, LLM-based methods, independent learning methods (e.g., MA2C: 595.30 seconds trip time, 383.86 seconds trip delay, 0.65 veh/s completion rate), and parameter-sharing methods (e.g., HeteroLight: 443.30 seconds trip time, 228.67 seconds trip delay, 1.03 veh/s completion rate). 
Among parameter-sharing methods, CoLight achieves the most competitive results, mainly due to its GAT-based feature extraction mechanism, which effectively aggregates critical information from neighboring intersections. 
Its completion rate (1.01 veh/s), trip time (425.31 seconds), and trip delay (216.29 seconds) are second only to those of LATS.
It is important to consider trip time and delay alongside the completion rate, as lower completion rates can lead to overestimated trip times and delays. LATS's high completion rate validates its efficient trip time and delay metrics. 

Similarly, we observe consistent results in the more complex heterogeneous Monaco network.
LATS achieves the lowest average queue length, the highest average speed, and a relatively low intersection delay, indicating its efficiency in reducing congestion and optimizing traffic flow for complex heterogeneous intersections.
It is worth noting that the Fixed-Time and AttendLight record slightly lower average intersection delay than LATS. 
This metric measures the mean \textbf{continuous waiting time} of vehicles, which resets each time a vehicle starts moving. 
Consequently, methods that switch phases more frequently tend to achieve lower delay values, even though such behavior may lead to less stable traffic progression. 
Therefore, this single metric does not fully capture the overall effectiveness of a control strategy and should be interpreted together with other indicators such as queue length, completion rate, and trip time/delay.
Additionally, LATS records the lowest trip time of 358.41 seconds and trip delay of 230.38 seconds while maintaining a high completion rate of 0.39 veh/s, significantly outperforming other methods.
Notably, the results of pure-LLM-based methods show that they cannot effectively understand and handle complex traffic conditions due to their lack of training on these specific control tasks.

Overall, LATS achieves significant improvements on most key metrics in both homogeneous Grid 5$\times$5 and heterogeneous Monaco networks. 
This highlights the robustness and generalizability of our LLM-assisted teacher-student MARL framework in managing diverse and dynamic urban traffic conditions.
In the \href{https://www.dropbox.com/scl/fi/zf1c7p81ua8cc6x6vzssm/Supplementary_Materials_T-ITS_LATS.pdf?rlkey=zfa2elenxz383ien27h63alja&st=adtp54jm&dl=0}{supplementary materials}, we provide plots of six traffic metrics changing over simulation steps during testing, which offer deeper insights than the averages and variances in the table. 
By showing how the metrics evolve at each step, these plots reveal the dynamic performance of different methods under varying traffic conditions throughout the simulation, enabling a more detailed analysis of their operational behavior.
Regarding inference speed, LATS takes less than $10^{-4}$ seconds per step per agent during execution process, while pure-LLM-based methods take about $6.5$ seconds. 
This indicates that the proposed plug-and-play teacher-student module can significantly reduce inference time.

To provide a more comprehensive evaluation of network-level performance, we introduce an extra evaluation metric, Average Trip Duration (ATD), which extends the conventional evaluation metrics adopted from MA2C~\cite{chu2019multi} by accounting for both unfinished trips and vehicles delayed before departure. 
Specifically, ATD measures the overall trip duration by combining the average travel time of departed vehicles, their departure delay, and the waiting time of vehicles that never departed. 
This formulation provides a fairer and more complete assessment of overall control performance (detailed definition is provided in the \href{https://www.dropbox.com/scl/fi/zf1c7p81ua8cc6x6vzssm/Supplementary_Materials_T-ITS_LATS.pdf?rlkey=zfa2elenxz383ien27h63alja&st=adtp54jm&dl=0}{supplementary materials}). 
Based on this refined metric, we evaluate both conventional and representative RL-based ATSC methods. 
The results in Table~\ref{exp:exp5} show that LATS consistently achieves the lowest ATD on both the homogeneous Grid 5×5 and heterogeneous Monaco networks, outperforming the best baseline by 8.51\% and 11.77\%, respectively.
This indicates that even when unfinished and delayed-departure vehicles are considered, LATS remains the best-performing method, demonstrating the superiority of our approach in optimizing network-wide traffic flow.

\begin{table*}
\caption{Optimization performance of various LATS variants and AttendLight variants in the ablation study.}
\label{exp:exp3}
\resizebox{\textwidth}{!}{%
\begin{tabular}{c|cccccc}
\hline &
  \begin{tabular}[c]{@{}c@{}}Queue Length $\downarrow$ \\ (vehs)\end{tabular} &
  \begin{tabular}[c]{@{}c@{}}Speed $\uparrow$ \\ (m/s)\end{tabular} &
  \begin{tabular}[c]{@{}c@{}}Intersections Delay $\downarrow$ \\ (sec)\end{tabular} &
  \begin{tabular}[c]{@{}c@{}}Completion Rate $\uparrow$ \\ (vehs/s)\end{tabular} &
  \begin{tabular}[c]{@{}c@{}}Trip Time $\downarrow$ \\ (sec)\end{tabular} &
  \begin{tabular}[c]{@{}c@{}}Trip Delay $\downarrow$ \\ (sec)\end{tabular} \\ \hline
\multicolumn{7}{c}{Homogeneous Grid 5$\times$5 network} \\ \hline
LATS &
  \textbf{1.26 (1.04)} &
  \textbf{4.73 (2.39)} &
  \textbf{13.60 (9.91)} &
  \textbf{1.03 (0.58)} &
  \textbf{389.80 (288.08)} &
  \textbf{182.72 (218.25)} \\
LATS w/o T &
  \underline{1.69 (1.20)} &
  \underline{4.44 (2.41)} &
  \underline{25.74 (18.89)} &
  \underline{1.01 (0.46)} &
  \underline{451.00 (365.52)} &
  230.90 (291.86) \\
LATS w/o S &
  1.78 (1.17) &
  3.89 (2.19) &
  26.57 (20.32) &
  0.98 (0.47) &
  499.38 (372.50) &
  \underline{227.58 (272.80)} \\
LATS w/o TS &
  4.30 (2.44) &
  1.92 (1.44) &
  53.36 (34.60) &
  0.60 (0.33) &
  935.52 (621.47) &
  687.05 (563.01) \\ \hline \hline
AttendLight w/ TS &
  \textbf{2.83 (1.84)} &
  \textbf{3.52 (2.41)} &
  \textbf{49.18 (36.01)} &
  \textbf{0.94 (0.41)} &
  \textbf{550.86 (450.81)} &
  \textbf{339.75 (387.31)} \\
AttendLight &
  3.37 (2.17) &
  3.22 (2.55) &
  62.89 (46.97) &
  0.85 (0.37) &
  569.80 (491.76) &
  358.32 (426.50) \\ \hline
\multicolumn{7}{c}{Heterogeneous Monaco network} \\ \hline
LATS &
  \textbf{1.05 (0.66)} &
  \textbf{3.87 (4.05)} &
  76.48 (52.33) &
  \textbf{0.39 (0.24)} &
  \textbf{358.41 (431.60)} &
  \textbf{230.38 (408.35)} \\
LATS w/o T &
  1.31 (0.83) &
  3.54 (3.99) &
  95.64 (65.00) &
  0.33 (0.22) &
  \underline{387.03 (492.42)} &
  \underline{259.12 (466.62)} \\
LATS w/o S &
  \underline{1.24 (0.86)} &
  \underline{3.66 (3.79)} &
  \textbf{62.93 (47.69)} &
  \underline{0.38 (0.25)} &
  443.12 (508.24) &
  303.91 (476.08) \\
LATS w/o TS &
  1.40 (0.78) &
  2.37 (2.29) &
  \underline{73.70 (50.90)} &
  0.34 (0.18) &
  777.14 (678.98) &
  627.99 (670.14) \\ \hline \hline
AttendLight w/ TS &
  1.67 (1.28) &
  \textbf{3.53 (3.83)} &
  69.96 (57.64) &
  0.32 (0.27) &
  \textbf{379.69 (465.65)} &
  \textbf{252.16 (438.55)} \\
AttendLight &
  \textbf{1.46 (1.05)} &
  3.38 (3.58) &
  \textbf{69.41 (56.94)} &
  \textbf{0.37 (0.25)} &
  484.66 (503.37) &
  346.26 (475.15) \\ \hline
\end{tabular}}
\end{table*}

\subsection{Zero-Shot Transferability (Q2)}
To evaluate the transferability and generalizability of different RL methods in diverse scenarios, we conduct experiments where models trained on the heterogeneous Monaco network are tested on the homogeneous Grid 5$\times$5 network with different traffic demands (low, medium, and high).
The results, as shown in Table~\ref{exp:exp2}, highlight LATS's superior transferability, which consistently achieves lower average queue length and trip delay compared to other advanced RL-based general TSC methods such as HeteroLight and AttendLight under varying traffic demands. 
Especially in high traffic demand scenarios, LATS achieves a trip time of 681.23 seconds and a trip delay of 482.87 seconds, surpassing the second-best AttendLight's 910.05 seconds and 714.63 seconds.
While AttendLight records a lower intersection delay under medium and high demands, this is due to the frequent phase switching that resets vehicles' continuous waiting times rather than indicating better overall performance.
This indicates that LATS not only performs well in the environment it is trained but also adapts effectively to new and unseen scenarios. 

Additionally, we perform a zero-shot transfer experiment where the policy trained on the homogeneous Grid 5×5 network is directly tested on the heterogeneous Monaco network.
As shown in Table~\ref{exp:c1_exp2_grid_to_monaco}, LATS consistently achieves superior performance across several key metrics, including queue length, completion rate, trip time, and trip delay, outperforming other advanced RL-based general TSC methods such as HeteroLight and AttendLight.
We exclude IPPO from this comparison because it fails to learn an effective policy on the Grid 5×5 network, which leads to unstable or near-random behavior in the zero-shot transfer setting and prevents a fair comparison with other methods.
While all methods exhibit performance degradation when transferred to the Monaco network, this may result from the more uniform and less diverse training data in the homogeneous Grid 5×5 network, which makes the learned policies more sensitive to distribution shifts.
Despite this, LATS still achieves better overall traffic performance.
This demonstrates that LATS not only performs well in the network it is trained on but also generalizes effectively to more complex and heterogeneous environments.

Overall, the results demonstrate that LATS effectively enhances the transferability of pure-RL-based methods by leveraging LLMs' semantic understanding and rich representational capabilities. 
The LLM-assisted teacher-student framework allows RL agents to learn from detailed, context-aware semantic representations generated by the embedding LLM, which captures the underlying traffic dynamics in a multi-modal manner compared to traditional representations. This deeper understanding of the traffic environment enables RL agents to generalize better across different scenarios, improving their ability to handle unseen scenarios.  Moreover, incorporating the knowledge and insights from embedding LLMs helps RL agents develop more robust strategies that do not overly rely on the specific features of the training scenarios. Instead, the agents learn to recognize broader patterns and relationships within the traffic data, which makes it easier to transfer these strategies to different contexts, thereby enhancing overall performance across diverse and complex traffic scenarios.

\begin{figure*}
    \centering
    \includegraphics[width=0.9\linewidth]{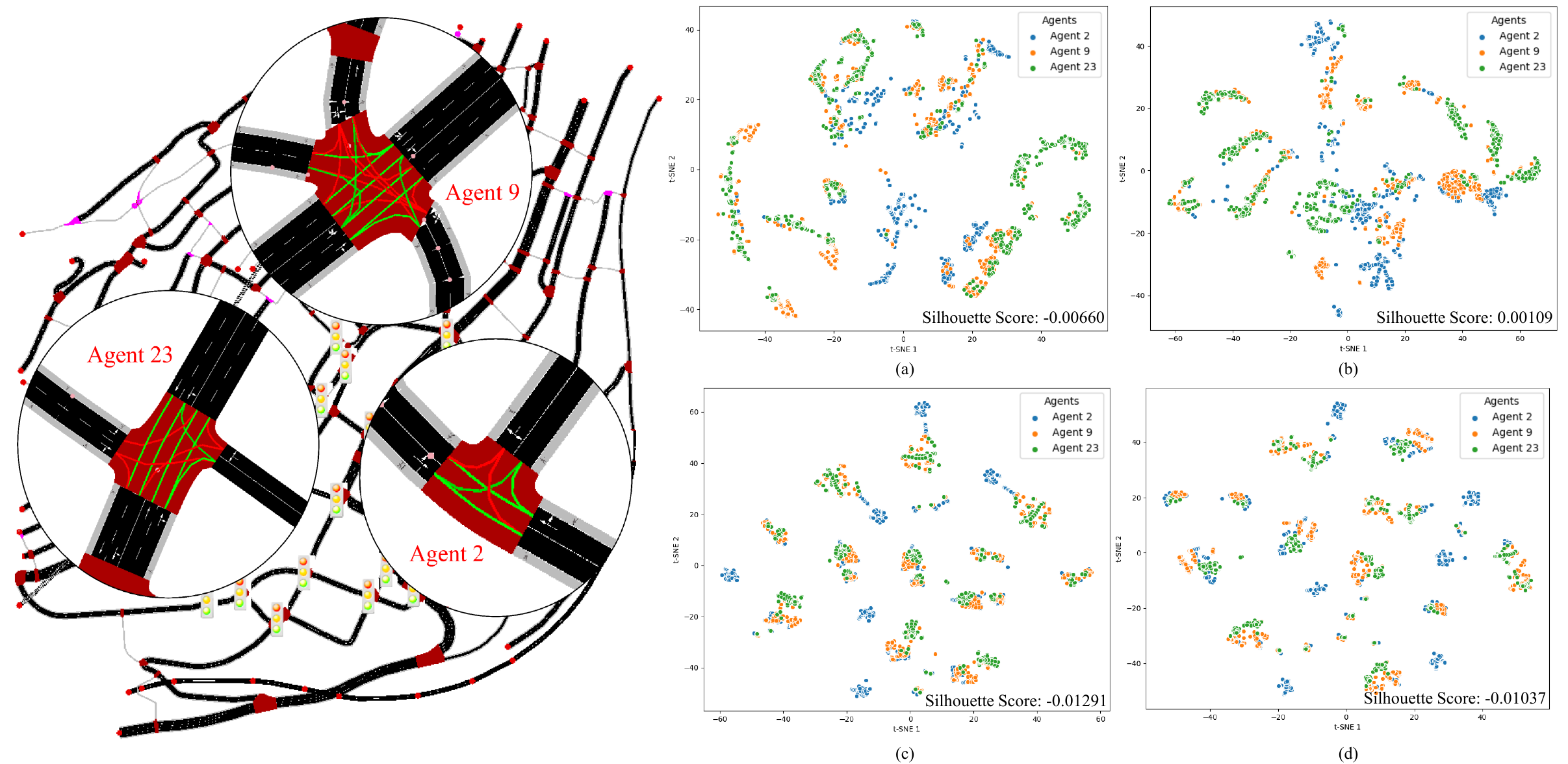}
    \caption{
    Illustration of the three selected intersections (agent 2, agent 9, and agent 23) with diverse topology structures from the heterogeneous Monaco traffic network used for t-SNE visualizations (left).
    2D t-SNE visualization of phase features for three selected intersections, comparing LATS and IPPO (LATS w/o TS) during training: (a) LATS after 800 episodes, (b) LATS after 1600 episodes, (c) IPPO after 800 episodes, and (d) IPPO after 1600 episodes (right).
    }
    \label{fig:tsne_results}
\end{figure*}

\subsection{Ablation Study (Q3)}
\label{sec:ablation_study}
\subsubsection{Ablation Study on LATS Components}
We conduct ablation study to evaluate the effectiveness of the LLM-assisted teacher-student module by comparing it with its different variants and integrating it with another general TSC method, AttendLight~\cite{oroojlooy2020attendlight}.
The results in Table~\ref{exp:exp3} show that removing the LLM-assisted teacher-student module from LATS leads to distinct performance degradation.
Notably, the variant LATS w/o S encodes the LLMs' semantic features into the latent space. It directly integrates these latent variables into the RL decision-making process, without relying on the feature distillation. 
The results show that using embedding LLMs as powerful semantic feature extractors and incorporating these multi-modal features in the RL process significantly enhances the performance of parameter-sharing RL methods. 
This improvement is likely due to the rich and diverse semantic features provided by embedding LLMs, which enhances the model's representational ability and promotes more diverse control strategies. 
However, it still relies on the LLM during execution, which slows down its inference speed and poses challenges for real-world deployment. 
In contrast, our teacher-student learning module balances accuracy and inference speed, effectively emulating the LLMs' semantic features through feature distillation in the latent space while maintaining a fast speed suitable for real-world TSC applications.

Furthermore, the results of the LATS w/o T variant indicate that even without guidance from the embedding LLMs, directly encoding topology and traffic state vectors into the latent space and integrating the resulting latent vectors into the RL decision-making process still improves the performance of parameter-sharing RL models. 
This may be because encoding topology and traffic state vectors in the latent space transforms complex traffic information into more expressive latent variables. 
This enables the model to capture crucial patterns and relationships within the traffic data, thus leading to better decision-making in complex traffic environments.
However, the improvements of LATS w/o T remain limited compared to the complete LATS framework, highlighting that while the VAE module provides useful compression and regularization, the semantic guidance from the embedding LLM that enhances the models' representational capability through teacher–student distillation is the primary driver of the performance gains.
Overall, LATS achieves the best overall performance, particularly on the key metrics of average queue length, trip completion rate, average trip time, and trip delay.

To further illustrate how our LLM-assisted teacher-student framework enhances the parameter-sharing RL methods, we visualize the phase features learned by LATS and IPPO (i.e., LATS w/o TS) across three selected agents in the Monaco network using the t-SNE method~\cite{van2008visualizing}.  
For each agent, we visualize the features of the first three phases within one episode (totaling 720 phase features including padding). These three agents are illustrated on the left side of Fig.~\ref{fig:tsne_results}. Here agent 2 is a three-road intersection with six traffic movements and two traffic phases, agent 9 is a four-road intersection with eleven traffic movements and four traffic phases, while agent 23 is a five-road intersection with eighteen traffic movements and five traffic phases.
As shown on the right side of Fig.~\ref{fig:tsne_results}, LATS shows a stronger capability in distinguishing phase features across different agents, with less overlap compared to IPPO, even after 800 and 1600 episodes of training. This reduced overlap suggests that LATS has a stronger representational ability, as it more effectively captures the features representing the unique topology and traffic dynamics of different agents. However, IPPO exhibits more overlap, indicating that it is less effective at distinguishing between different traffic scenarios and capturing each agent's specific characteristics.

In addition to visualizing the phase features, we introduce the Silhouette Score~\cite{rousseeuw1987silhouettes} to quantify how well-separated the different clusters of agent phase features are within the feature space.
A higher Silhouette Score means better separation between clusters, while a lower or negative score means there is significant overlap. The Silhouette Scores are calculated as: $-6.60 \times 10^{-3}$ for LATS after 800 episodes, $1.09 \times 10^{-3}$ after 1600 episodes, $-1.29 \times 10^{-2}$ for IPPO after 800 episodes, and $-1.04 \times 10^{-2}$ after 1600 episodes. 
These quantitative results indicate that our LATS effectively captures and differentiates features of different traffic scenarios, thereby enhancing the model's representational capacity.
Overall, these results show that integrating semantic features from LLMs and applying knowledge distillation in the latent space significantly enhances LATS' ability to represent and adapt to complex traffic scenarios. This leads to greater generalization and robustness, showing the effectiveness of LATS in improving model representation for more efficient control strategies.

\begin{table*}[t!]
\caption{
Ablation study of different LATS prompt variants on the homogeneous \textit{Grid 5×5} network and the heterogeneous \textit{Monaco} network.
}
\label{exp:c1_exp4_prompt_ablation}
\resizebox{\textwidth}{!}{%
\begin{tabular}{c|cccccc}
\hline 
& \begin{tabular}[c]{@{}c@{}}Queue Length $\downarrow$\\ (vehs)\end{tabular} &
  \begin{tabular}[c]{@{}c@{}}Speed $\uparrow$ \\(m/s)\end{tabular} &
  \begin{tabular}[c]{@{}c@{}}Intersections Delay $\downarrow$\\ (sec)\end{tabular} &
  \begin{tabular}[c]{@{}c@{}}Completion Rate $\uparrow$ \\(vehs/sec)\end{tabular} &
  \begin{tabular}[c]{@{}c@{}}Trip Time $\downarrow$ \\ (sec)\end{tabular} &
  \begin{tabular}[c]{@{}c@{}}Trip Delay $\downarrow$ \\ (sec)\end{tabular} \\ 
\hline
\multicolumn{7}{c}{Homogeneous Grid 5$\times$5 network with 25 signalized intersections} \\ \hline
LATS w/ Full &
    \textbf{1.26 (1.04)} &
    \underline{4.73 (2.39)} &
    \textbf{13.60 (9.91)} &
    \textbf{1.03 (0.58)} &
    \textbf{389.80 (288.08)} &
    \textbf{182.72 (218.25)} \\
LATS w/ Topo &
    1.50 (1.19) &
    4.58 (2.45) &
    17.45 (11.88) &
    \textbf{1.03 (0.54)} &
    431.24 (343.67) &
    216.30 (267.82) \\
LATS w/ Dyna &
    \underline{1.38 (1.10)} &
    \textbf{4.74 (2.33)} &
    \underline{16.12 (11.31)} &
    \textbf{1.03 (0.54)} &
    \underline{404.81 (329.44)} &
    \underline{200.00 (258.96)} \\ \hline \hline
LATS w/ Jina &
    \textbf{1.26 (1.04)} &
    4.73 (2.39) &
    \textbf{13.60 (9.91)} &
    \textbf{1.03 (0.58)} &
    \textbf{389.80 (288.08)} &
    \textbf{182.72 (218.25)} \\
LATS w/ BGE &
    1.33 (1.06) &
    \textbf{4.85 (2.26)} &
    14.09 (10.22) &
    \textbf{1.03 (0.54)} &
    405.75 (303.99) &
    191.51 (233.49) \\  \hline
\multicolumn{7}{c}{Heterogeneous Monaco network with 28 signalized intersections} \\ \hline
LATS w/ Full &
    \textbf{1.05 (0.66)} &
    \textbf{3.87 (4.05)} &
    \textbf{76.48 (52.33)} &
    \textbf{0.39 (0.24)} &
    \textbf{358.41 (431.60)} &
    \textbf{230.38 (408.35)} \\
LATS w/ Topo &
    1.27 (0.82) &
    \underline{3.71 (4.06)} &
    97.82 (72.29) &
    0.35 (0.23) &
    377.06 (504.81) &
    249.76 (479.38) \\
LATS w/ Dyna &
    \underline{1.26 (0.82)} &
    3.66 (4.02) &
    \underline{82.69 (55.29)} &
    0.35 (0.23) &
    \underline{387.05 (531.72)} &
    \underline{262.36 (506.62)} \\ \hline \hline
LATS w/ Jina &
    \textbf{1.05 (0.66)} &
    \textbf{3.87 (4.05)} &
    76.48 (52.33) &
    \textbf{0.39 (0.24)} &
    \textbf{358.41 (431.60)} &
    \textbf{230.38 (408.35)} \\ 
LATS w/ BGE &
    1.18 (0.75) &
    3.77 (4.07) &
    \textbf{76.03 (49.39)} &
    0.36 (0.23) &
    371.73 (509.63) &
    246.63 (486.92) \\ \hline
\end{tabular}}
\end{table*}

\subsubsection{Ablation Study on Prompt Design}
To evaluate the effectiveness of our prompt design for embedding generation in ATSC, we conduct ablation experiments comparing three prompt variants: the full prompt (LATS w/ Full), the topology-only prompt (LATS w/ Topo), and the dynamics-only prompt (LATS w/ Dyna), as shown in Table~\ref{exp:c1_exp4_prompt_ablation}.
The results show that, in both homogeneous and heterogeneous networks, the complete prompt variant (LATS w/ Full) consistently achieves the best performance across almost all metrics, while removing either prompt component leads to performance degradation.
In the homogeneous Grid 5×5 network, the variant using only topological information (LATS w/ Topo) performs worse than the one using only dynamic traffic information (LATS w/ Dyna), likely because all intersections share the same topology and the main variation among prompts arises from traffic dynamics.
Consequently, excluding topological information has a smaller impact, whereas removing dynamic information results in a more significant performance drop.
We hypothesize that prompts containing richer information that captures the unique characteristics of each intersection, particularly its traffic dynamics, generate more distinctive semantic embeddings.
These embeddings, when incorporated into the RL model, enhance its representational ability to better distinguish between intersections, thereby improving policy diversity, generalization, and overall performance.
In contrast, in the heterogeneous Monaco network, where intersections differ substantially in topology, both topological and dynamic components are crucial for accurately representing intersection semantics.
The results support this observation: removing either component leads to a performance decline, indicating that both types of information play important roles in forming a comprehensive semantic representation at intersections.

\subsubsection{Ablation Study on Embedding LLMs} 
To evaluate the impact of different embedding LLMs on overall performance, we compare two variants of the LATS framework: one using \textit{jina-embeddings-v2-small-en} (LATS w/ Jina) and the other using \textit{bge-small-en-v1.5}\footnote{\url{https://huggingface.co/BAAI/bge-small-en}}
 (LATS w/ BGE).
The BGE model~\cite{bge_embedding} is an open-source embedding LLM designed for general-purpose text representation.
It is pretrained with a masked autoencoding objective and further refined through contrastive learning on large-scale paired text data, making it a representative and strong baseline in text embedding tasks.
As shown in Table~\ref{exp:exp5}, on both the homogeneous Grid 5×5 and heterogeneous Monaco networks, the variant using \textit{jina-embeddings-v2-small-en} achieves slightly better overall performance across most key evaluation metrics. 
This advantage can be attributed to the \textit{Jina}-based model’s training on more diverse and large-scale datasets, as well as its higher embedding dimensionality of 512 compared with 384 in the BGE model. 
Together, these factors enable it to capture more expressive and fine-grained semantic representations of intersection topologies and traffic dynamics, which in turn contribute to improved downstream RL performance.
When compared with other advanced ATSC baselines in Table~\ref{exp:exp1}, both variants consistently outperform existing methods, indicating that the proposed LATS framework remains effective and robust regardless of the specific embedding LLM adopted. 
Overall, these results demonstrate that incorporating semantic representations from embedding LLMs enhances both the representational capacity and control performance of RL models in ATSC tasks, highlighting the effectiveness of the proposed LLM-assisted framework.

\section{Limitations and Future Work}
While our method effectively enhances semantic representation at intersections and improves the RL performance for ATSC, several limitations remain to be addressed.

\textbf{LLM dependency and training efficiency:}
Although the proposed LATS framework introduces a teacher–student distillation paradigm that allows the final policy to operate independently from the LLM during practical deployment, the training process still relies on LLM-generated embeddings. 
This dependency introduces a relatively high computational cost, which substantially increases the overall training time.
In future work, we will explore more efficient learning schemes to reduce this overhead, such as enabling the student model to learn effectively from fewer environment interactions or adopting cross-network joint training to improve sample efficiency and accelerate convergence across diverse traffic networks.

\textbf{Generalization and representation limitations:}
While LATS enhances the representational capacity of RL models by integrating multi-modal semantic features, it does not fully address the common out-of-distribution issue in RL~\cite{packer2018assessing}, where performance may degrade when encountering unseen traffic demands or network topologies. 
Moreover, the current framework relies on fixed prompt templates and textual inputs, which may limit the expressiveness and variability of the learned semantics.
In addition, our method does not explicitly consider more realistic traffic conditions, including weather changes, traffic accidents, and sudden demand surges, mainly because most open-source traffic simulators are not designed to reproduce such complex and unpredictable situations.
Future work will focus on developing advanced prompt enhancement strategies for more informative and adaptive semantic representations.
We also plan to extend the framework beyond textual inputs by incorporating visual and auditory modalities, thereby broadening the applicability of the approach.

\textbf{Lack of explicit inter-agent communication:}
This work focuses primarily on improving intersection-level phase representations rather than modeling explicit communication or cooperation among agents. 
As a result, the framework currently lacks mechanisms for semantic interaction or message exchange between intersections.
Future research will extend LATS toward communication-aware coordination by leveraging LLMs to generate interpretable and context-aware messages among agents, promoting cooperative decision-making and improving network-level traffic control performance.

\section{Conclusion}
In this work, we present LATS, a novel ATSC framework that integrates LLMs with MARL, aiming to leverage LLMs' strong prior knowledge and inductive capabilities to enhance the RL decision-making process by providing precise semantic understanding rather than creative content generation.
Specifically, we introduce a plug-and-play teacher-student learning module, where a trained embedding LLM serves as the teacher to generate semantic features based on traffic prompts. While a much simpler student neural network learns to emulate these features through knowledge distillation. 
Such integration effectively combines the strong inductive abilities and rich semantic understanding of LLMs, with the robustness and fast inference of MARL, thus leading to more efficient, robust, and generalizable control strategies.
Extensive experiments across diverse traffic datasets demonstrate the superiority of LATS over other advanced MARL and LLM-only approaches. 
We also show our method can easily integrate with other RL-based general TSC frameworks to improve their network-wide traffic optimization performance.
These results indicate the effectiveness of LATS and its potential to open new avenues for integrating RL and LLMs in practical TSC applications.
 
\bibliographystyle{ieeetr}
\bibliography{sample}

\end{document}